\documentclass{article}
\usepackage{manuscript,times}
\usepackage{graphicx}
\usepackage{caption}
\usepackage{booktabs}

\usepackage{amsmath,amsfonts,amssymb,bm}

\usepackage{hyperref}
\hypersetup{
  colorlinks=false,
  pdfborder={0 0 1},
  citebordercolor={0 1 0},
  linkbordercolor={1 0 0},
  urlbordercolor={1 1 1}
}
\usepackage{url}

\title{Map2SLAM: From Streaming Map Prediction to Real-Time SLAM}

\author{%
  \parbox[t]{\dimexpr\textwidth-2\tabcolsep\relax}{%
    \raggedright\normalfont\normalsize
    Jian Gao$^{1,2}$\hspace{5pt}Lin-Zhuo Chen$^{1,2}$\hspace{5pt}%
    Yihang Chen$^{3,2}$\hspace{5pt}Jiagang Chen$^{4,2}$\hspace{5pt}%
    Jiacheng Dong$^{5,2}$\hspace{5pt}Xun Cao$^{1}$ \\[3pt]
    Nan Xue$^{2}$\hspace{5pt}%
    Yinghao Xu$^{3,}$\thanks{Corresponding authors.}\hspace{5pt}Yao Yao$^{1,\dagger}$ \\[7pt]
    {\footnotesize
    $^{1}$Nanjing University\hspace{7pt}$^{2}$Robbyant, Ant Group\hspace{7pt}%
    $^{3}$HKUST\hspace{7pt}$^{4}$Wuhan University\hspace{7pt}$^{5}$Zhejiang University\endgraf}
  }%
}

\usepackage{xcolor}

\begin{document}

\setcounter{footnote}{1}
\maketitle
\vspace{-1em} 
\begin{center}
  \centering
  \captionsetup{type=figure}
  \includegraphics[pagebox=cropbox,width=\textwidth]{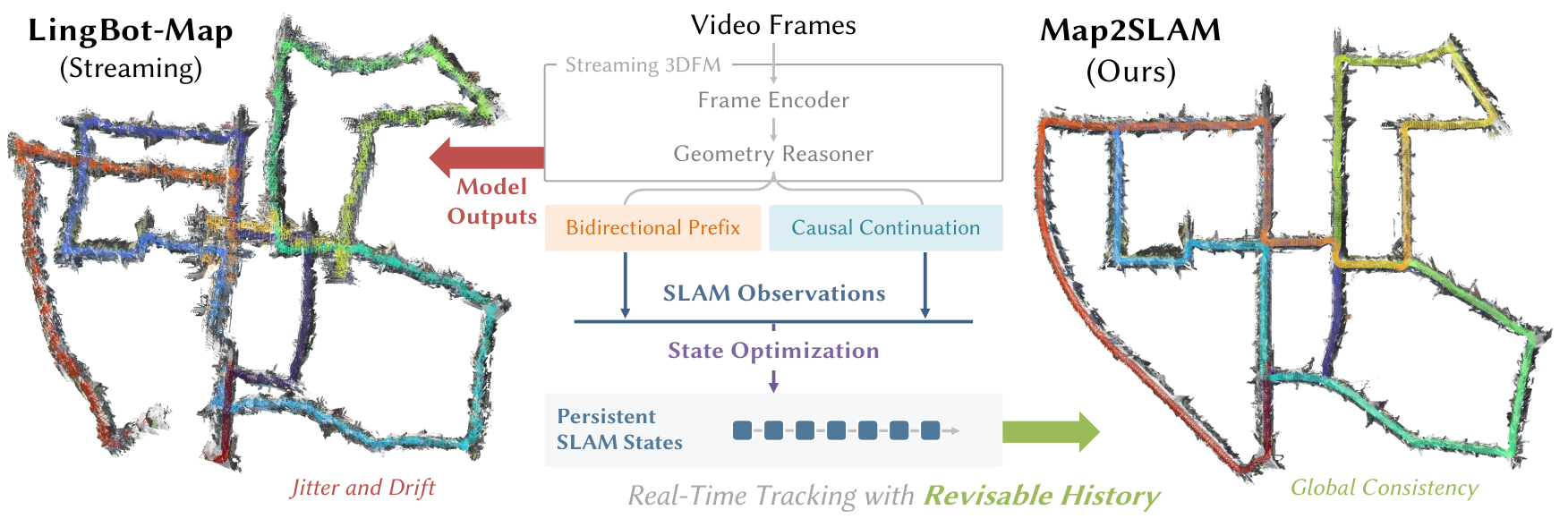}
  \captionsetup{skip=6pt}
  \caption{\textbf{From Streaming Map Prediction to SLAM.} Model outputs from LingBot-Map exhibit jitter and drift (left). Rather than treating these predictions as final estimates, Map2SLAM uses them as observations of persistent SLAM states (center). The Geometry Reasoner follows one prefix-causal attention architecture: selected history forms a fresh \emph{Bidirectional Prefix} for re-observation, while \emph{Causal Continuation} observes incoming frames. Both reuse features from a single Frame Encoder pass. These operations enable real-time tracking with a revisable history, yielding a globally consistent reconstruction (right).}
  \label{fig:teaser}
\end{center}

\begin{abstract}

\medskip
Streaming 3D foundation models (3DFMs) estimate camera poses and dense geometry
online, but they fall short of SLAM: their predictions are fixed once emitted,
so loop closures cannot correct accumulated drift. With such models already
providing strong geometric capability, a central question for SLAM becomes how
to schedule their inference and how to keep its results as revisable state.
We present Map2SLAM, which keeps LingBot-Map frozen, trains only a lightweight
retrieval head, and builds SLAM on the model's inference interface, turning its
predictions into observations of an external persistent SLAM state. The
front-end tracks every incoming frame with the model's causal streaming
inference and decides which frames are written to the persistent state as
keyframes. The back-end asynchronously decides when and which past frames to
re-observe, re-running the model's joint inference over them with fresh
context; verified loop observations correct the trajectory online, and segment
observations refine dense geometry after the sequence. All invocations
share a single frame encoder pass per image, whose cached features also drive
loop retrieval, so frequent re-invocation incurs no additional encoding cost. 
Map2SLAM achieves state-of-the-art performance on KITTI, VBR, and Oxford Spires, 
reducing mean ATE by over 40\% compared with the best uncalibrated baselines, 
while maintaining real-time tracking at over 28 FPS at a resolution of \(518\times154\) on an NVIDIA RTX 5090D GPU.
Please visit our project page at \href{https://robbyant-research.github.io/Map2SLAM/}{\textcolor{magenta}{https://robbyant-research.github.io/Map2SLAM/}}.

\end{abstract}
\section{Introduction}
\label{sec:introduction}

Feed-forward 3D foundation models (3DFMs) infer camera poses and dense scene
geometry directly from unposed images~\citep{wang2024dust3r,wang2025vggt}.
Streaming 3DFMs extend this capability to online video, producing a geometric
prediction for each incoming frame~\citep{zhuo2026streaming,chen2026geometric}.
However, these predictions are fixed once emitted: drift accumulates over long
sequences, and even when the camera revisits a known place, earlier estimates
cannot be corrected (Figure~\ref{fig:teaser}, left). Current streaming 3DFMs
therefore cannot yet replace SLAM.

Prior work has combined 3DFMs with SLAM in several ways: aligning chunk-wise
predictions~\citep{deng2026vggt-long,lee2026swiftvggt}, using a 3DFM as the
front-end of a classical SLAM pipeline~\citep{murai2025mast3r,
maggio2025vggt-slam,maggio2026vggt-slam2}, or attaching loop closure to a
streaming model~\citep{cheng2026horizonstream,han2026revisiting}. These methods
invoke the model in a fixed pattern, over predefined chunks, frame pairs, or a
single causal pass, rather than scheduling its inference. Given the strong geometric
capability of current models, we argue that the key to building SLAM lies in
how to schedule model inference and how to keep its results as revisable
state. Just as LLM agents rely on a harness to schedule a fixed model, a 3DFM
needs a similar mechanism to decide which frames enter its context, when to
re-observe history, and which state each prediction updates.

Following this principle, we build Map2SLAM upon
LingBot-Map~\citep{chen2026geometric} with a framework that schedules its
inference and turns its predictions into persistent SLAM state, yielding
accurate real-time SLAM. The
prefix-causal structure of LingBot-Map supports two
inference modes: \emph{Bidirectional Prefix} jointly reasons over multiple
frames, and \emph{Causal Continuation} infers frame by frame autoregressively
over long videos. Map2SLAM assigns the two modes to the two ends of the system:
the front-end processes the live stream with Causal Continuation, and the
back-end re-observes history with Bidirectional Prefix. The front-end decides
which frames enter the context and are written to the state. To avoid the
trajectory jitter caused by an ever-growing context, it tracks within bounded
windows and treats each Causal Continuation step as a \emph{Keyframe Probe}:
based on the projective motion induced by the predicted geometry, an accepted
frame is committed as a keyframe without repeated inference, and a rejected
frame has its KV update rolled back.

Windowed tracking still accumulates drift over long sequences. The back-end
revises the state asynchronously, deciding when and which past frames to
re-observe. Whereas LingBot-Map uses Bidirectional Prefix only for
initialization, the back-end re-invokes it with fresh context. \emph{Loop
Retrieval} proposes revisit candidates from cached features. Because appearance
matches do not guarantee geometric consistency, \emph{Pairwise Geometry}
verifies each candidate with a two-view prefix, and verified constraints enter
\emph{Incremental Pose Graph Optimization} (iPGO), which corrects the
trajectory online. The back-end also re-observes sequential and loop segments
with multi-view prefixes, whose results feed \emph{Dense Geometric Refinement}
after the sequence ends. All invocations share a single \emph{Frame Encoder}
pass per image: both the downstream Geometry Reasoner and the retrieval head
reuse its cached features, so frequent re-invocation incurs no additional
encoding cost.

We evaluate Map2SLAM on KITTI~\citep{geiger2012we}, VBR~\citep{brizi2024vbr},
and Oxford Spires~\citep{tao2026oxford}. Map2SLAM achieves state-of-the-art performance on KITTI, VBR, and Oxford Spires, reducing mean ATE by more than 40\% relative to the best-performing uncalibrated baselines, while running in real time at over 28 FPS with an input resolution of \(518\times154\) on an NVIDIA RTX 5090D GPU.

Our contributions are threefold:
\begin{itemize}
    \item We propose a design principle for building SLAM on streaming 3DFMs:
    given their strong geometric capability, the key lies in scheduling model
    inference and managing the state formed by its results.

    \item We present Map2SLAM, which realizes this principle on a frozen
    LingBot-Map: a causal front-end writes the state through commit-or-rollback
    Keyframe Probes, and an asynchronous back-end revises it by re-invoking
    Bidirectional Prefix with fresh context, with all invocations sharing one
    Frame Encoder pass per image.

    \item Map2SLAM achieves the lowest mean ATE among uncalibrated methods on
    KITTI, VBR, and Oxford Spires, 42.7--48.3\% below the second-best results,
    while tracking at over 28 FPS on an NVIDIA RTX 5090D GPU.
\end{itemize}

\section{Related Work}
\label{sec:related}

\paragraph{Long-Sequence Visual SLAM}\par
Visual SLAM has evolved from feature-based systems~\citep{klein2007parallel,mur2015orb}, through direct methods~\citep{engel2014lsd,engel2018direct}, to learned geometric frontends~\citep{teed2021droid,teed2023deep}. Long-sequence SLAM, however, requires more than estimating motion as frames arrive. It must retain past estimates in its state so that a loop can introduce long-range constraints and revise the trajectory~\citep{mur2015orb,campos2021orb,engel2014lsd,gao2018ldso,lipson2024deep}.

\paragraph{3D Foundation Models}\par
Recent 3D foundation models (3DFMs) infer camera and scene geometry in a feed-forward pass over image collections~\citep{wang2024dust3r,zhang2025flare,wang2025vggt,wang2026vggt,fang2026dens3r,gao2026more}. Streaming 3DFMs extend this inference to video frames as they arrive~\citep{zhuo2026streaming,chen2026geometric,cheng2026horizonstream,han2026revisiting}, enabling geometric prediction over increasingly long sequences.
In particular, \mbox{LingBot-Map~\citep{chen2026geometric}} first processes an initial set of frames jointly to establish scene context, and then incrementally incorporates subsequent frames in an autoregressive manner.

\paragraph{Long-Sequence 3D Reconstruction}\par
Recent work extends 3DFMs to long sequences by composing local predictions into globally consistent reconstructions~\citep{deng2026vggt-long,lee2026swiftvggt,zhang2026vggtalign,ding2026laser,zhang2026talo,xiong2026vggt}.
Related efforts increasingly integrate 3DFM predictions with incremental estimation, loop closure, and global optimization~\citep{murai2025mast3r,maggio2025vggt-slam,maggio2026vggt-slam2,mandal2026vggt,jeon2026aim,lee2026unisim,fang2026slamformer}.
Together, these developments are carrying long-sequence reconstruction beyond a purely offline setting and progressively closer to the system-level concerns of SLAM.

\section{Method}
\label{sec:method}

Map2SLAM turns streaming geometric predictions into observations that update
persistent trajectory and dense-geometry states. It builds on
LingBot-Map~\citep{chen2026geometric}, which first processes an initial frame
set jointly with bidirectional attention for coordinate grounding and then
handles later frames causally, one at a time, for online inference. We call
these executions \emph{Bidirectional Prefix} and \emph{Causal Continuation};
together, they form a \emph{prefix-causal architecture}~\citep{dong2019unified}.
Map2SLAM uses Causal Continuation for live tracking and runs Bidirectional
Prefix again on selected historical frames to produce fresh geometric
observations for optimization.

\subsection{System Overview}

\begin{figure}[t]
  \centering
  \includegraphics[width=\linewidth]{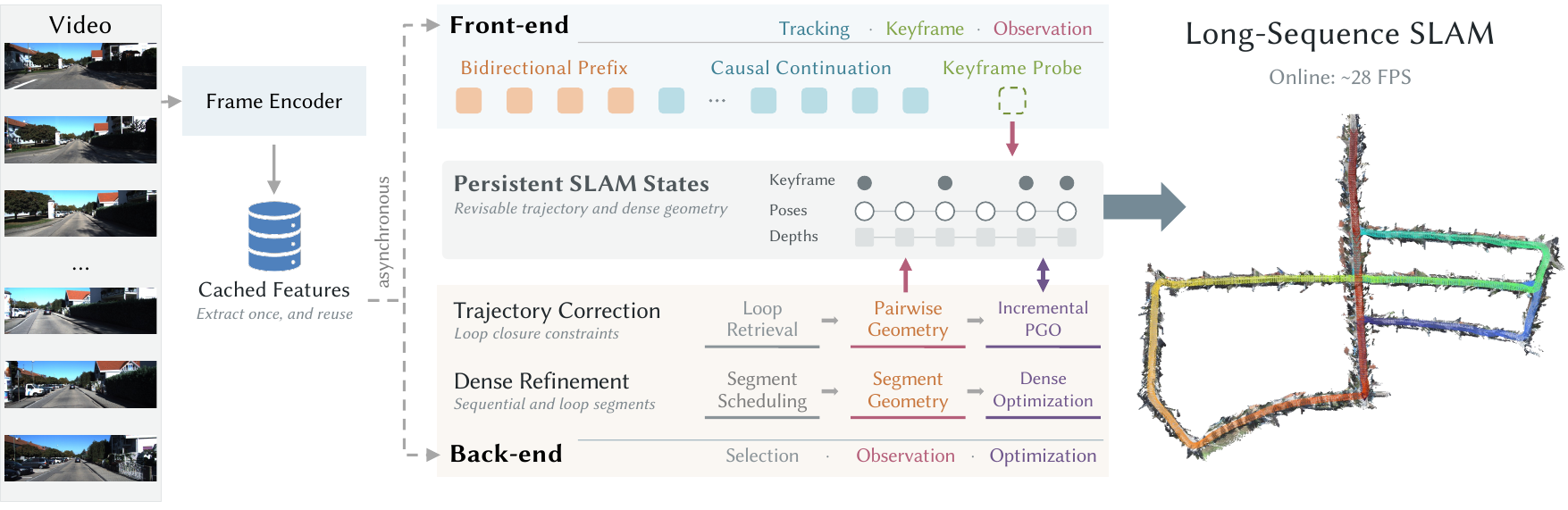}
  \captionsetup{skip=4pt}
  \caption{\textbf{System overview.} The Frame Encoder processes each image once, and cached outputs are reused throughout. The Front-end initializes each tracking window with a Bidirectional Prefix, tracks through Causal Continuation, and admits keyframes through Keyframe Probe. Each Back-end path performs selection, fresh-context re-observation, and optimization. Loop Retrieval, Pairwise Geometry, and iPGO correct the trajectory online; Segment Scheduling, Segment Geometry, and Dense Optimization refine keyframe poses and depths after the sequence. Final keyframe corrections propagate to non-keyframes.}
  \label{fig:pipeline}
\end{figure}

In our Map2SLAM system, Front-end tracking and Back-end re-observation run asynchronously on separate
GPUs. The \emph{Frame Encoder} processes each image once: the Front-end consumes
its geometric features immediately, while Back-end queries reuse cached outputs
from the same pass. Within each tracking window, the Front-end maintains one
active \emph{Geometry Reasoner} context; each Back-end geometric query starts
from a fresh context. Figure~\ref{fig:pipeline} summarizes this architecture.

\subsection{Front-end Tracking}

Extending LingBot-Map's streaming context over long sequences leads to
trajectory \textit{jitter}. We therefore track in bounded windows and
reinitialize the active \emph{Geometry Reasoner} context at each boundary. This
bounds the context and stabilizes local tracking, but introduces drift between
windows.

To initialize the first window, we process its first \(N_p\) frames as
a Bidirectional Prefix, establishing the initial tracking context. For each
subsequent frame, one Causal Continuation step tentatively extends the active
context to form a Keyframe Probe, yielding the pose, depth, and intrinsics used
to measure projective motion. Let \(k(i)\) denote the latest keyframe before
frame \(I_i\). We measure the projective motion induced by the probe as
\begin{equation}
    m_i =
    \frac{1}{|\Omega_{k(i)}|}
    \sum_{\mathbf{u}\in\Omega_{k(i)}}
    \left\|
    \pi\!\left(
        K_i T_{i\leftarrow k(i)}
        \pi^{-1}\!\left(
            \mathbf{u}, D_{k(i)}(\mathbf{u}), K_{k(i)}
        \right)
    \right)
    - \mathbf{u}
    \right\|_2 ,
    \label{eq:projective-motion}
\end{equation}
where \(\mathbf{u}\) denotes a pixel coordinate in keyframe \(k(i)\);
\(T_i\), \(K_i\), and \(D_i\) are the predicted camera pose, intrinsics, and
depth, respectively; \(\pi\) and \(\pi^{-1}\) denote perspective projection
and back-projection; \(\Omega_{k(i)}\) contains valid pixels of keyframe
\(k(i)\); and \(T_{i\leftarrow k(i)}=T_i^{-1}T_{k(i)}\). If \(m_i>\tau\), the
front-end accepts the candidate frame as a keyframe and commits its KV cache
update; otherwise, it rejects the frame and rolls back the update.

When the context reaches its budget \(N_b\), accepting a frame triggers context
reinitialization rather than a commit. The front-end clears the KV cache and
forms the successor Bidirectional Prefix from the accepted frame and the last
\(N_p-1\) keyframes. This prefix has its own similarity gauge, which we align
to the existing trajectory using the replayed keyframes: their depth ratios
determine the scale \(s_w\), while their poses determine the rotation \(R_w\)
and translation \(t_w\). Each local prediction
\((\widetilde R_i,\widetilde t_i,\widetilde D_i)\) in the successor context is
then mapped as
\begin{equation}
    R_i = R_w\widetilde R_i,\qquad
    t_i = s_wR_w\widetilde t_i+t_w,\qquad
    D_i = s_w\widetilde D_i.
    \label{eq:window-alignment}
\end{equation}
This alignment preserves cross-window continuity for subsequent causal outputs.

For each accepted keyframe, the Front-end records its pose relative to the
preceding accepted keyframe as a sequential observation. For each non-keyframe,
it retains the causal pose relative to the latest keyframe for final trajectory
recovery. Online Trajectory Correction combines the sequential observations
with verified loop constraints to correct accumulated long-range drift.

\subsection{Back-end Optimization}

The Back-end comprises two paths with the same three stages: \textit{selection},
\textit{geometric re-observation}, and \textit{optimization}. Online Trajectory Correction
updates the trajectory while tracking continues; Dense Geometric Refinement
jointly refines keyframe poses and depths after the sequence.

\subsubsection{Online Trajectory Correction}
\label{sec:global-trajectory-correction}
\paragraph{Loop Retrieval}
Although LingBot-Map is trained to predict dense geometry, we find that
intermediate tokens from its Frame Encoder remain effective for visual place
recognition. We add a lightweight SALAD aggregation
head~\citep{izquierdo2024optimal} to these cached tokens while keeping
the encoder frozen. The head produces a global descriptor for each accepted
keyframe, and descriptor similarity proposes historical loop candidates for
Pairwise Geometry. 

To control the number of redundant loop candidates, each descriptor becomes
searchable only after \(Q=64\) subsequent keyframes have been accepted, and
each query retains up to \(K=5\) loop candidates with SALAD similarity of at
least \(\tau_s=0.40\). Once a pair is verified, we suppress further candidates
between
the accepted-keyframe neighborhoods of radius \(r=5\) around its two
endpoints.

\begin{figure}[t]
  \centering
  \includegraphics[width=\linewidth]{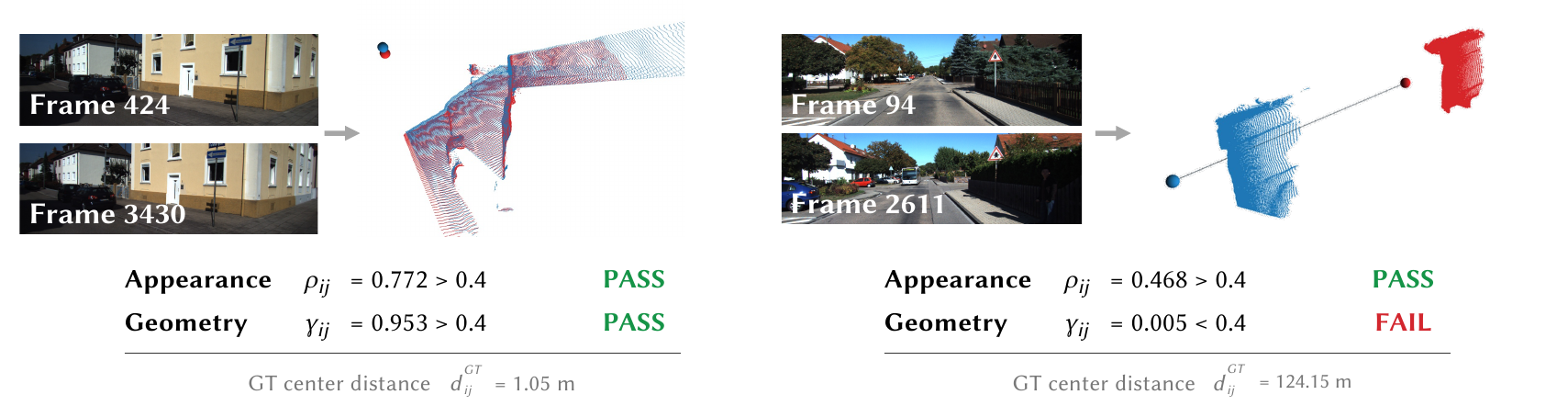}
  {\small
  \begin{tabular}{@{}p{0.5\linewidth}@{}p{0.5\linewidth}@{}}
    \centering (a) Accepted after geometric verification. &
    \centering (b) Rejected by geometric verification. \tabularnewline
  \end{tabular}}
  \captionsetup{skip=4pt}
  \caption{\textbf{Loop Retrieval followed by Pairwise Geometry}
  ($\tau_{\rho}=\tau_{\gamma}=0.4$). Both pairs pass appearance retrieval; fresh two-view geometric re-observation accepts (a) and rejects (b). Ground-truth camera-center distance is shown only for reference and is not used by the method.}
  \label{fig:loop}
\end{figure}

\paragraph{Pairwise Geometry}
Pairwise Geometry consists of \emph{relative-geometry estimation} followed by
\emph{geometric consistency verification}. For each loop candidate \((i,j)\),
a fresh two-view Bidirectional Prefix over cached frame features jointly
predicts the two camera poses, depth maps, and intrinsics. We then verify the
predicted geometry through bidirectional depth consistency:
\begin{equation}
\gamma_{i\rightarrow j}
=
\frac{1}{|\Omega_i|}
\sum_{\mathbf{p}\in\Omega_i}
\mathbf{1}\!\left[e_{i\rightarrow j}(\mathbf{p}) < \epsilon\right],
\qquad
e_{i\rightarrow j}(\mathbf{p})
=
\frac{
\left|\hat d_{i\rightarrow j}(\mathbf{p})
-d_j\!\left(\pi_{i\rightarrow j}(\mathbf{p})\right)\right|
}{
d_j\!\left(\pi_{i\rightarrow j}(\mathbf{p})\right)
}.
\end{equation}
Here, \(\Omega_i\) contains source pixels with valid depth, while
\(\pi_{i\rightarrow j}(\mathbf{p})\) and
\(\hat d_{i\rightarrow j}(\mathbf{p})\) denote the projected location and
depth of \(\mathbf{p}\) in view \(j\), respectively. A projected depth is
consistent when its relative error is below \(\epsilon=0.10\); invalid
projections are counted as inconsistent. We use
\(\gamma_{ij}=\min\{\gamma_{i\rightarrow j},\gamma_{j\rightarrow i}\}\) as the
pairwise consistency score. If
\(\gamma_{ij}\geq\tau_g\), with \(\tau_g=0.40\), the candidate yields a
verified loop constraint for iPGO; otherwise, it is discarded, as shown in
Figure~\ref{fig:loop}.

\paragraph{Incremental Pose Graph Optimization}
At time \(t\), iPGO operates on the ordered accepted keyframes
\(\mathcal{K}_t=(k_0,\ldots,k_{N_t-1})\) and the loop candidates
\(\mathcal{C}_t\) proposed up to that time. Causal Tracking and Pairwise
Geometry provide the sequential and verified-loop constraint sets
\begin{equation}
    \mathcal{E}_{\mathrm{seq},t}
    =
    \left\{
        (k_{n-1},k_n)
    \right\}_{n=1}^{N_t-1},
    \qquad
    \mathcal{E}_{\mathrm{loop},t}^{\mathrm{ver}}
    =
    \left\{
        (i,j)\in\mathcal{C}_t
        \;\middle|\;
        \gamma_{ij}\geq\tau_g
    \right\}.
    \label{eq:ipgo-constraint-sets}
\end{equation}
iPGO optimizes
\(\mathcal S_t=\{\mathbf S_i\}_{i\in\mathcal K_t}\), where
\(\mathbf S_i\in\mathrm{Sim}(3)\). Each constraint \((i,j)\) supplies a
relative measurement \(\hat{\mathbf S}_{ij}\in\mathrm{Sim}(3)\). The
\(\mathrm{Sim}(3)\) representation allows joint correction of scale, rotation,
and translation; fixing \(\mathbf S_{k_0}\) to its initial value removes the
global gauge freedom.

Because Pairwise Geometry verifies each candidate independently, verified loop
constraints may still conflict with the current trajectory. We jointly
evaluate them against the sequential constraints using
GNC-TLS~\citep{yang2020graduated} in riSAM's incremental
scheme~\citep{mcgann2023robust}, treating the sequential constraints as known
inliers. Each verified loop receives a robust weight
\(w_{ij,t}\in[0,1]\), and the active set is
\(\mathcal{E}_{\mathrm{loop},t}^{\mathrm{act}}
=\{(i,j)\in\mathcal{E}_{\mathrm{loop},t}^{\mathrm{ver}}
\mid w_{ij,t}\geq0.5\}\). As new constraints arrive, the selector updates
these weights without changing the loop measurements, allowing loops to be
activated, withdrawn, or reactivated.

The sequential and active loop constraints enter a stateful iSAM2
graph~\citep{kaess2012isam2}, which estimates the trajectory by minimizing
\begin{equation}
    \mathcal S_t^\star
    =
    \arg\min_{\mathcal S_t}
    \sum_{(i,j)\in
    \mathcal E_{\mathrm{seq},t}
    \cup
    \mathcal E_{\mathrm{loop},t}^{\mathrm{act}}}
    \rho_{\mathrm H}\!\left(
        d_{\mathrm{Sim}(3)}
        \bigl(\hat{\mathbf S}_{ij},\mathbf S_{ij}(\mathcal S_t)\bigr)
    \right).
    \label{eq:ipgo-objective}
\end{equation}
Here, \(\mathbf S_{ij}(\mathcal S_t)\) is the relative transformation implied
by the current keyframe states, \(d_{\mathrm{Sim}(3)}\) compares it with the
measurement \(\hat{\mathbf S}_{ij}\), and \(\rho_{\mathrm H}\) is the Huber
loss. As keyframes and active-set changes arrive, iSAM2 adds new states and
sequential constraints and adds or removes loop constraints, revising the
trajectory without rebuilding the graph.

\subsubsection{Dense Geometric Refinement}

\paragraph{Segment Scheduling}
We first group accepted keyframes for historical re-observation into two types
of segments. For \textit{Sequential Segments}, we slide a window of \(N_s=24\)
keyframes with an overlap of \(N_o=6\) over the accepted-keyframe order. For a loop pair
\((i,j)\), \(i<j\), we form a \textit{Loop Segment} by concatenating
\(N_s/2\) keyframes ending at \(k_j\) with \(N_s/2\) keyframes starting at
\(k_i\). This ordering places the verified revisit at one adjacent seam in a
virtual capture sequence. Sequential Segments become available upon window
completion and Loop Segments upon loop verification.

\paragraph{Segment Geometry}
For each available segment, we run a fresh Bidirectional Prefix
over its cached features. Unlike Causal Continuation, every
keyframe can attend to all other keyframes in the segment, providing full
multi-view context. Each execution outputs local poses, per-pixel depths, intrinsics, 
and confidence maps. These service as one segment observation. Reusing the cached 
features avoids another Frame Encoder pass, and these independent executions can 
overlap with Front-end Causal Tracking.

\paragraph{Dense Optimization}
After the input sequence ends and iPGO produces its final active set, we retain
all Sequential Segment observations and the Loop Segment observations
associated with this set, denoted by \(\mathcal O\). We initialize one shared
pose \(T_i\) per keyframe from the final iPGO trajectory and optimize these
poses together with one log-depth field \(z_i\) per keyframe and shared camera
parameters \(\boldsymbol{\alpha}\). For each retained segment \(s\), we fit a
\(\mathrm{Sim}(3)\) alignment \(A_s\) from its prefix-local poses to their iPGO
counterparts and keep \(A_s\) fixed. The shared 3D point at pixel \(\mathbf u\)
is
\(\mathbf X_i(\mathbf u)=T_i\pi^{-1}(\mathbf u,\exp z_i(\mathbf u),
K(\boldsymbol{\alpha}))\).

Following confidence-weighted 3D pointmap
alignment~\citep{wang2024dust3r}, we optimize
\begin{equation}
\begin{aligned}
\min_{\{T_i,z_i\},\,\boldsymbol{\alpha}}\quad
&\sum_{\substack{s\in\mathcal O,\,i\in\mathcal K_s\\
\mathbf u\in\Omega_{si}}}
w_{si}(\mathbf u)\,
\rho_{\delta}\!\left(
\left\|\mathbf r_{si}(\mathbf u)\right\|_2
\right),\\
\mathbf r_{si}(\mathbf u)
&=
\frac{
A_s^{-1}\mathbf X_i(\mathbf u)
-\widetilde{\mathbf X}_{si}(\mathbf u)
}{
\sigma_r\widetilde{\ell}_{si}(\mathbf u)
}.
\end{aligned}
\label{eq:dense-pose-depth}
\end{equation}
Here, \(\mathcal K_s\) contains the keyframes in segment \(s\), and
\(\Omega_{si}\) contains valid positive-depth predictions with confidence at
least one. The weight \(w_{si}\) is the predicted log confidence normalized by
the number of retained segments covering keyframe \(i\). We set
\(\sigma_r=0.1\) and use the Huber loss with \(\delta=1.5\). We fix the first
keyframe pose, solve the remaining variables with Levenberg--Marquardt, and
eliminate the per-pixel log depths at each iteration using an exact Schur
complement. The optimized state yields refined keyframe poses, dense depths,
and shared camera intrinsics.

\subsection{Pose Correction Propagation}
The back-end refines only keyframes, while non-keyframes retain their Causal Tracking poses 
relative to the preceding keyframe. The change from each keyframe's causal pose to its final pose 
defines a Sim(3) correction. We interpolate these corrections between adjacent keyframes and apply 
them to intermediate non-keyframes, recovering the full trajectory without additional model inference.

\section{Experiments}
\label{sec:experiments}

\subsection{Benchmark Evaluation}
We evaluate trajectory accuracy on KITTI Odometry~\citep{geiger2012we},
VBR~\citep{brizi2024vbr}, and Oxford Spires~\citep{tao2026oxford}, using ATE
after a global \(\mathrm{Sim}(3)\) alignment. We report per-sequence ATE and
the mean across sequences, grouping methods into calibrated and uncalibrated
settings. Methods with frequent tracking failures are not reported on the
corresponding benchmarks: ORB-SLAM3 is reported only on KITTI, while
MASt3R-SLAM~\citep{murai2025mast3r} is omitted from all three. VGGT-SLAM 2.0
is included for reference only, as it reports poses at keyframes.

\begin{table*}[t]
  \centering
  \captionsetup{skip=2pt}
  \caption{Absolute trajectory error (ATE, m) on KITTI~\citep{geiger2012we}.
  ``--'' denotes tracking failure. LC, Dense Geo., and Frame-wise denote loop
  closure, dense geometry, and per-frame tracking, respectively.
  LingBot-Map(S/W) denote the streaming/window modes. $^\dagger$ VGGT-SLAM 2.0
  uses native keyframes only and is excluded from ranking. The best and
  second-best results in each section are shown in bold and underlined,
  respectively.}
  \label{tab:kitti-results}
  \scriptsize
  \setlength{\tabcolsep}{2.0pt}
  \renewcommand{\arraystretch}{1.18}
  \newcommand{\seqmeta}[1]{{\color{gray}\itshape\fontsize{4.2}{4.6}\selectfont #1}}
  \newcommand{\seqhead}[3]{%
    \shortstack{\rule{0pt}{2.4ex}\textbf{#1}\\[0.45ex]
    \seqmeta{#2}\\[0.10ex]
    \seqmeta{#3}\rule[-0.9ex]{0pt}{0.9ex}}}
  \resizebox{\linewidth}{!}{%
  \begin{tabular}{@{}l|c|cc|ccccccccccc|c@{}}
    \toprule
    \raisebox{1.7ex}{\textbf{Method}} & \raisebox{1.7ex}{\textbf{LC}} &
    \raisebox{0.7ex}{\shortstack{\textbf{Dense}\\\textbf{Geo.}}} &
    \raisebox{0.7ex}{\shortstack{\textbf{Frame-}\\\textbf{wise}}} &
    \seqhead{00}{4541\,f}{3.70\,km} &
    \seqhead{01}{1101\,f}{2.50\,km} &
    \seqhead{02}{4661\,f}{5.10\,km} &
    \seqhead{03}{801\,f}{0.60\,km} &
    \seqhead{04}{271\,f}{0.40\,km} &
    \seqhead{05}{2761\,f}{2.20\,km} &
    \seqhead{06}{1101\,f}{1.20\,km} &
    \seqhead{07}{1101\,f}{0.70\,km} &
    \seqhead{08}{4071\,f}{3.20\,km} &
    \seqhead{09}{1591\,f}{1.70\,km} &
    \seqhead{10}{1201\,f}{0.90\,km} &
    \raisebox{1.7ex}{\textbf{Avg. $\downarrow$}} \\
    \midrule
    \multicolumn{16}{@{}l}{\emph{Calibrated}} \\
    ORB-SLAM3~\citep{campos2021orb} & $\checkmark$ & & $\checkmark$
      & \textbf{7.77} & -- & \textbf{25.18} & \textbf{0.85} & \textbf{0.72} & \underline{6.03} & \underline{16.93} & \textbf{2.57} & \textbf{64.55} & \textbf{6.76} & \textbf{8.48} & -- \\
    DROID-SLAM~\citep{teed2021droid} & $\checkmark$ & $\checkmark$ & $\checkmark$
      & 110.19 & \underline{90.94} & 138.72 & 3.20 & 1.47 & 73.64 & 61.50 & \underline{16.09} & \underline{103.85} & 92.01 & 22.21 & \underline{64.89} \\
    DPV-SLAM++~\citep{lipson2024deep} & $\checkmark$ & & $\checkmark$
      & \underline{7.87} & \textbf{27.48} & \underline{71.75} & \underline{2.56} & \underline{1.03} & \textbf{5.92} & \textbf{12.49} & 18.00 & 113.94 & \underline{74.54} & \underline{13.65} & \textbf{31.75} \\
    \midrule
    \multicolumn{16}{@{}l}{\emph{Uncalibrated}} \\
    DROID-Calib~\citep{hagemann2023deep} & & $\checkmark$ & $\checkmark$
      & 47.65 & 34.21 & 93.95 & 5.78 & 2.39 & 39.74 & 11.29 & 18.62 & 76.32 & 68.26 & \underline{8.79} & 37.00 \\
    VGGT-Long~\citep{deng2026vggt-long} & $\checkmark$ & $\checkmark$ &
      & 10.83 & 66.41 & 55.69 & 5.57 & 3.95 & 10.13 & \underline{6.95} & 2.83 & 57.59 & 31.21 & 22.16 & 24.85 \\
    SwiftVGGT~\citep{lee2026swiftvggt} & $\checkmark$ & $\checkmark$ &
      & \underline{9.30} & 113.82 & 56.29 & 10.39 & 5.43 & 11.82 & 8.35 & 24.56 & 68.90 & 31.08 & 25.07 & 33.18 \\
    VGGT-SLAM 2.0$^\dagger$~\citep{maggio2026vggt-slam2} & $\checkmark$ & $\checkmark$ &
      & 104.44 & 151.66 & 146.87 & 12.92 & 2.92 & 10.36 & 28.58 & 18.25 & 76.39 & 28.84 & 24.04 & 55.02 \\
    LoGeR~\citep{zhang2026loger} & & $\checkmark$ &
      & 62.93 & 41.68 & \underline{39.18} & 4.87 & 1.82 & 41.09 & 13.89 & 16.18 & 26.43 & 22.82 & 8.81 & 25.43 \\
    LingBot-Map(S)~\citep{chen2026geometric} & & $\checkmark$ & $\checkmark$
      & 31.33 & 102.42 & 56.71 & 3.93 & 1.27 & 23.59 & 8.33 & 8.63 & \underline{19.09} & 21.50 & 12.52 & 26.30 \\
    LingBot-Map(W)~\citep{chen2026geometric} & & $\checkmark$ & $\checkmark$
      & 52.46 & 106.71 & 63.17 & \underline{3.29} & 2.06 & 30.39 & 8.39 & 9.69 & 50.95 & 54.84 & 8.99 & 35.54 \\
    HorizonStream~\citep{cheng2026horizonstream} & $\checkmark$ & $\checkmark$ & $\checkmark$
      & 15.38 & \underline{23.89} & 70.80 & 5.56 & \textbf{0.71} & 7.94 & 7.73 & 3.16 & 22.29 & 27.21 & 12.29 & 17.91 \\
    ABot-Recon~\citep{han2026revisiting} & $\checkmark$ & $\checkmark$ & $\checkmark$
      & 11.67 & \textbf{21.49} & 39.82 & 4.36 & 2.90 & \underline{5.91} & 8.65 & \textbf{1.66} & 22.39 & \underline{14.04} & 16.22 & \underline{13.56} \\
    \textbf{Ours} & $\checkmark$ & $\checkmark$ & $\checkmark$
      & \textbf{3.51} & 26.94 & \textbf{11.02} & \textbf{1.92} & \underline{1.02} & \textbf{3.40} & \textbf{2.06} & \underline{2.24} & \textbf{16.18} & \textbf{9.65} & \textbf{5.24} & \textbf{7.56} \\
    \bottomrule
  \end{tabular}}
\end{table*}

\begin{table*}[t]
  \centering
  \captionsetup{skip=2pt}
  \caption{Absolute trajectory error (ATE, m) on VBR~\citep{brizi2024vbr}.
  ``--'' denotes tracking failure. LC, Dense Geo., and Frame-wise denote loop
  closure, dense geometry, and per-frame tracking, respectively.
  LingBot-Map(S/W) denote the streaming/window modes. $^\dagger$ VGGT-SLAM 2.0
  uses native keyframes only and is excluded from ranking. The best and
  second-best results in each section are shown in bold and underlined,
  respectively.}
  \label{tab:vbr-results}
  \scriptsize
  \setlength{\tabcolsep}{2.0pt}
  \renewcommand{\arraystretch}{1.18}
  \newcommand{\vbrseqmeta}[1]{{\color{gray}\itshape\fontsize{4.2}{4.6}\selectfont #1}}
  \newcommand{\vbrseqhead}[3]{%
    \shortstack{\rule{0pt}{2.4ex}\makebox[5em][c]{\vphantom{Ag}\textbf{#1}}\\[0.45ex]
    \vbrseqmeta{#2}\\[0.10ex]
    \vbrseqmeta{#3}\rule[-0.9ex]{0pt}{0.9ex}}}
  \resizebox{\linewidth}{!}{%
  \begin{tabular}{@{}l|c|cc|ccccccc|c@{}}
    \toprule
    \raisebox{1.7ex}{\textbf{Method}} & \raisebox{1.7ex}{\textbf{LC}} &
    \raisebox{0.7ex}{\shortstack{\textbf{Dense}\\\textbf{Geo.}}} &
    \raisebox{0.7ex}{\shortstack{\textbf{Frame-}\\\textbf{wise}}} &
    \vbrseqhead{Colosseo-0}{8815\,f}{1.45\,km} &
    \vbrseqhead{Campus-0}{12042\,f}{2.73\,km} &
    \vbrseqhead{Campus-1}{11671\,f}{2.95\,km} &
    \vbrseqhead{Pincio-0}{11142\,f}{1.27\,km} &
    \vbrseqhead{Spagna-0}{14141\,f}{1.56\,km} &
    \vbrseqhead{Diag-0}{10021\,f}{1.02\,km} &
    \vbrseqhead{Ciampino-1}{18846\,f}{5.20\,km} &
    \raisebox{1.7ex}{\textbf{Avg. $\downarrow$}} \\
    \midrule
    \multicolumn{12}{@{}l}{\emph{Calibrated}} \\
    DROID-SLAM~\citep{teed2021droid} & $\checkmark$ & $\checkmark$ & $\checkmark$
      & \underline{76.66} & \underline{125.77} & \underline{83.87} & \underline{49.85} & \underline{59.61} & \underline{35.81} & \underline{145.28} & \underline{82.41} \\
    DPV-SLAM++~\citep{lipson2024deep} & $\checkmark$ & & $\checkmark$
      & \textbf{18.59} & \textbf{21.69} & \textbf{22.22} & \textbf{39.37} & \textbf{37.31} & \textbf{33.51} & \textbf{45.56} & \textbf{31.18} \\
    \midrule
    \multicolumn{12}{@{}l}{\emph{Uncalibrated}} \\
    DROID-Calib~\citep{hagemann2023deep} & & $\checkmark$ & $\checkmark$
      & 27.36 & 76.05 & 68.52 & 14.09 & 25.38 & -- & 124.98 & -- \\
    VGGT-Long~\citep{deng2026vggt-long} & $\checkmark$ & $\checkmark$ &
      & 44.13 & 132.70 & 115.62 & 64.29 & 59.05 & 33.66 & 179.41 & 89.84 \\
    SwiftVGGT~\citep{lee2026swiftvggt} & $\checkmark$ & $\checkmark$ &
      & 21.24 & 131.36 & 115.94 & 65.80 & 58.43 & 34.52 & 176.58 & 86.27 \\
    VGGT-SLAM 2.0$^\dagger$~\citep{maggio2026vggt-slam2} & $\checkmark$ & $\checkmark$ &
      & 18.75 & 26.88 & 22.90 & 33.25 & 10.17 & 6.47 & 49.80 & 24.03 \\
    LoGeR~\citep{zhang2026loger} & & $\checkmark$ &
      & 25.51 & 19.69 & 29.23 & 27.91 & 22.87 & 34.99 & 93.04 & 36.18 \\
    LingBot-Map(S)~\citep{chen2026geometric} & & $\checkmark$ & $\checkmark$
      & \underline{14.85} & 20.30 & 9.92 & 30.46 & 23.71 & 22.68 & 65.09 & 26.71 \\
    LingBot-Map(W)~\citep{chen2026geometric} & & $\checkmark$ & $\checkmark$
      & 86.86 & 123.09 & 104.64 & 63.40 & 56.68 & 23.68 & 82.20 & 77.22 \\
    HorizonStream~\citep{cheng2026horizonstream} & $\checkmark$ & $\checkmark$ & $\checkmark$
      & 25.00 & 28.79 & 10.54 & 15.49 & 21.62 & \underline{7.44} & 18.36 & 18.18 \\
    ABot-Recon~\citep{han2026revisiting} & $\checkmark$ & $\checkmark$ & $\checkmark$
      & 16.70 & \underline{8.00} & \underline{4.30} & \textbf{5.12} & \underline{17.91} & 9.69 & \underline{8.09} & \underline{9.97} \\
    \textbf{Ours} & $\checkmark$ & $\checkmark$ & $\checkmark$
      & \textbf{7.54} & \textbf{4.28} & \textbf{2.45} & \underline{6.57} & \textbf{3.38} & \textbf{4.67} & \textbf{7.18} & \textbf{5.15} \\
    \bottomrule
  \end{tabular}}
\end{table*}

\begin{figure*}[t]
  \centering
  \begingroup
  \newcommand{\trajectorypanel}[1]{%
    \includegraphics[width=0.326\textwidth]{#1}%
  }
  \setlength{\tabcolsep}{0pt}
  \begin{tabular}{@{}ccc@{}}
    \trajectorypanel{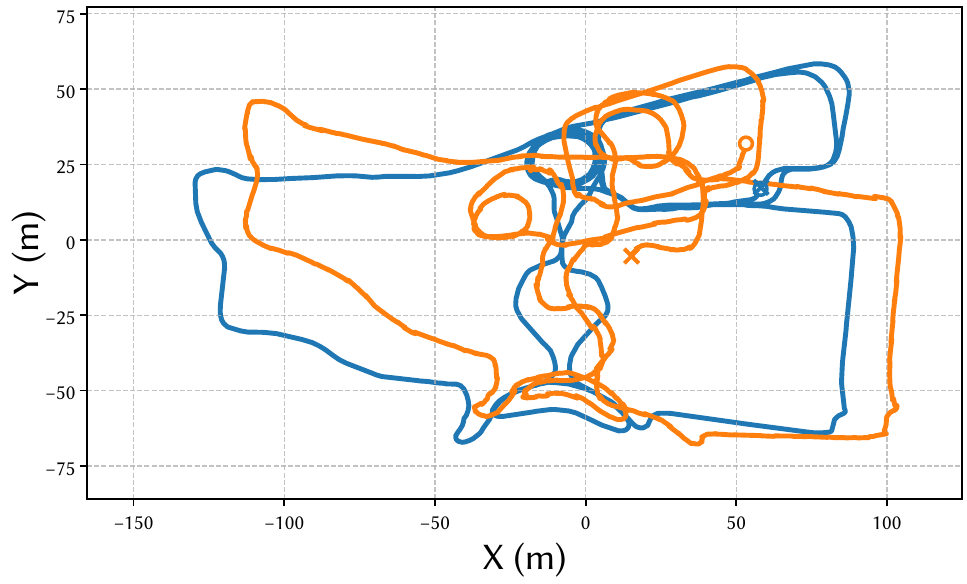} &
    \trajectorypanel{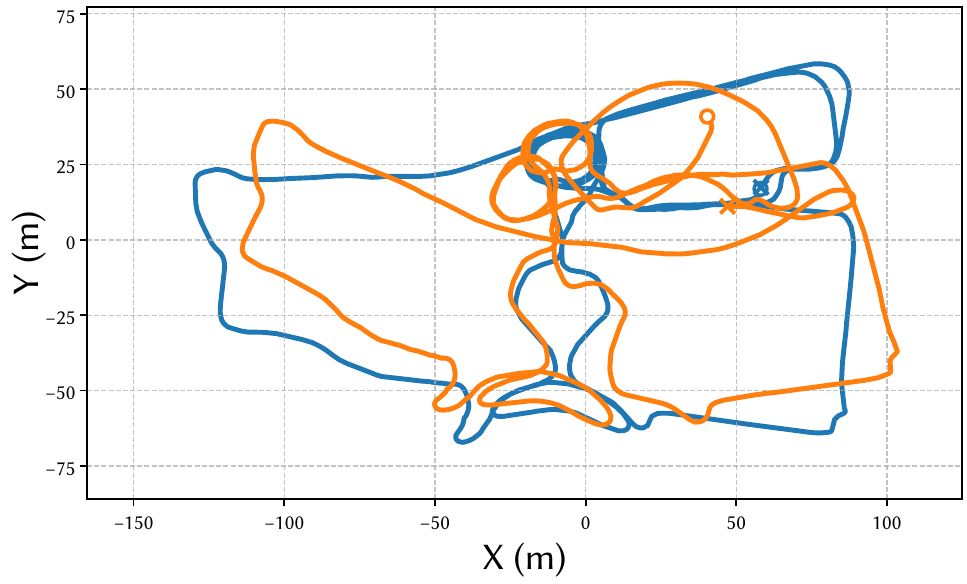} &
    \trajectorypanel{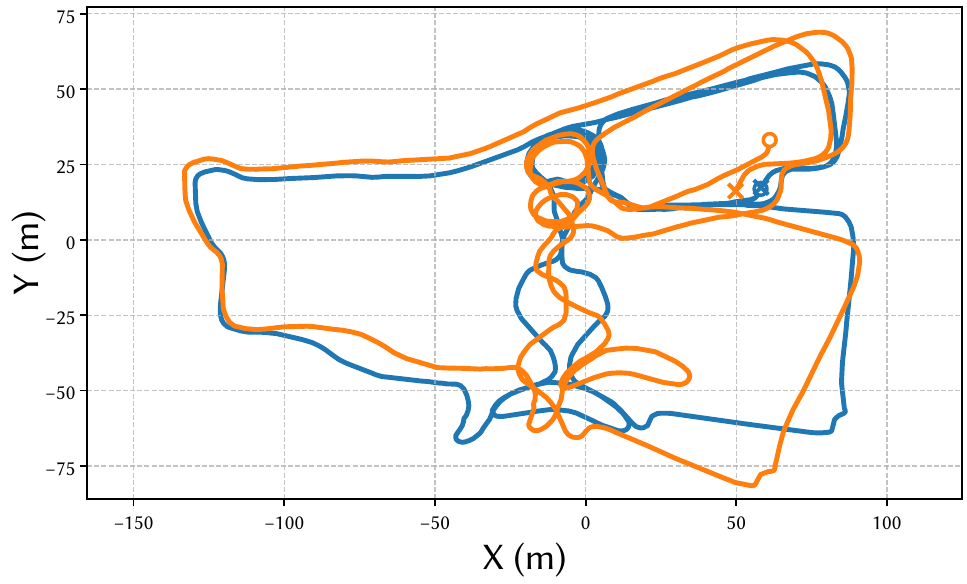} \\[-1pt]
    {\footnotesize (a) LoGeR} &
    {\footnotesize (b) HorizonStream + LC} &
    {\footnotesize (c) ABot-Recon + LC} \\[2pt]
    \trajectorypanel{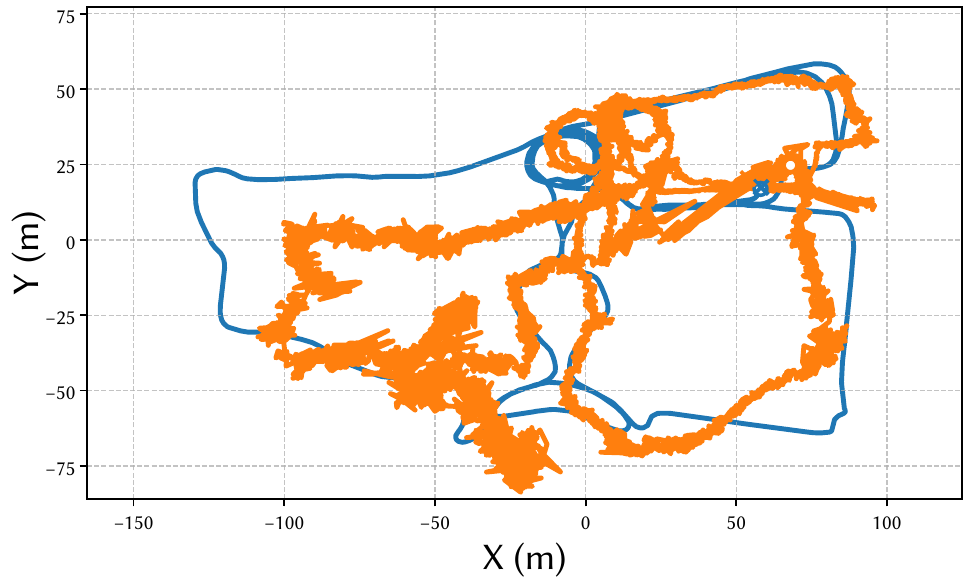} &
    \trajectorypanel{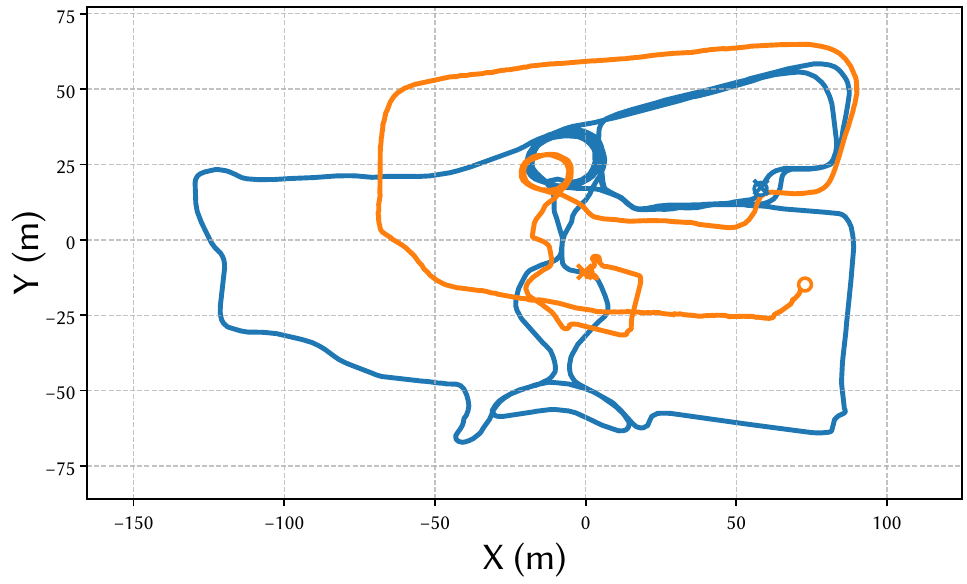} &
    \trajectorypanel{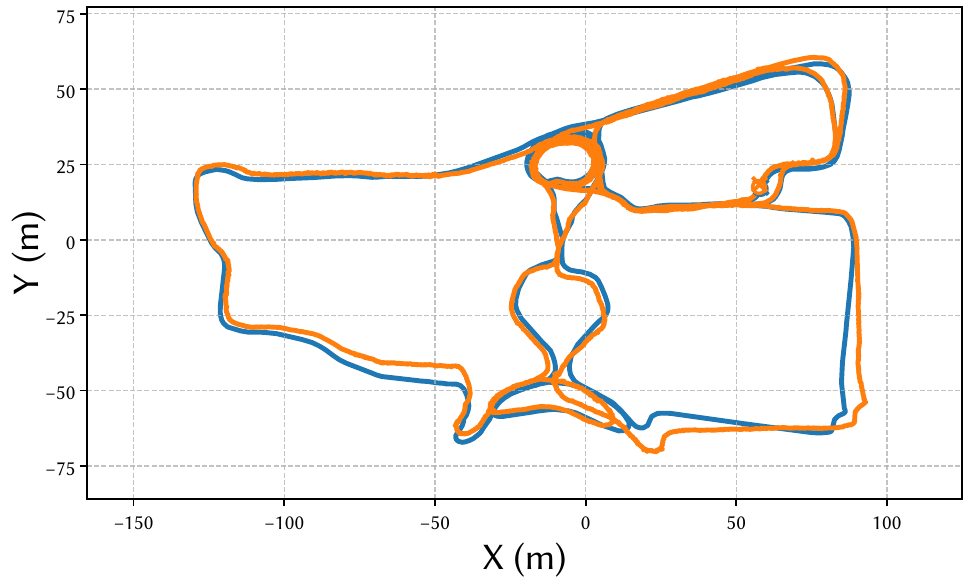} \\[-1pt]
    {\footnotesize (d) LingBot-Map (S)} &
    {\footnotesize (e) LingBot-Map (W)} &
    {\footnotesize (f) Ours}
  \end{tabular}
  \endgroup
  \captionsetup{skip=4pt}
  \caption{\textbf{Trajectory comparison on VBR \emph{Spagna-0}.}
  The sequence contains 14,141 frames and covers 1.56 km. Blue and orange
  denote the reference and estimated trajectories, respectively.}
  \label{fig:trajectory-comparison}
\end{figure*}

As shown in Figure~\ref{fig:trajectory-comparison}(d--f), LingBot-Map(S)
exhibits severe local jitter, whereas LingBot-Map(W) produces a smoother
trajectory but accumulates cross-window drift that distorts the global route.
Our system suppresses both effects and closely follows the ground truth.
This improvement is also reflected quantitatively, with substantially lower ATE on KITTI, VBR,
and Oxford Spires. These results suggest that model capability is only part 
of the equation; how its predictions are used matters too.

Against existing uncalibrated methods, our system achieves the lowest mean 
ATE on all three benchmarks, improving over ABot-Recon by more than \(40\%\) 
in each case. This advantage is consistent across sequences, with our 
system ranking first or second on 26 of the 28 sequences. Despite using 
no camera intrinsics, it also outperforms the best complete calibrated 
baseline on KITTI and VBR; on Oxford Spires, DPV-SLAM++ remains better 
at \(2.28\,\mathrm{m}\) versus \(3.07\,\mathrm{m}\).

\subsection{System Ablations}
\label{sec:system-ablations}

\newsavebox{\systemablationtablebox}
\newsavebox{\systemablationstagebox}
\newcommand{\systemablationstage}[2]{%
  \begingroup
  \sbox{\systemablationstagebox}{%
    \includegraphics[width=0.326\textwidth,pagebox=cropbox]{#2}%
  }%
  \usebox{\systemablationstagebox}%
  \llap{%
    \raisebox{\dimexpr\ht\systemablationstagebox-12pt\relax}{%
      \makebox[\wd\systemablationstagebox][l]{%
        \hspace{4pt}\scriptsize\bfseries #1%
      }%
    }%
  }%
  \endgroup
}

\begin{figure*}[t]
  \centering
  \sbox{\systemablationtablebox}{%
    \setlength{\tabcolsep}{3.5pt}%
    \renewcommand{\arraystretch}{1.45}%
    \resizebox{0.54\textwidth}{!}{%
      \begin{tabular}{@{}lcccc@{\hspace{7pt}}ccc@{}}
        \toprule
        \textbf{Configuration} &
        \shortstack{\textbf{Geometric}\\\textbf{Check}} &
        \textbf{GNC-TLS} &
        \textbf{Dense Opt.} &
        \shortstack{\textbf{Segment}\\$\mathbf{Sim(3)}$} &
        \shortstack{\textbf{ATE}\\\textbf{(m)} $\downarrow$} &
        \shortstack{\textbf{RPE$_t$}\\\textbf{(m)} $\downarrow$} &
        \shortstack{\textbf{RPE$_r$}\\\textbf{(deg)} $\downarrow$} \\
        \midrule
        Causal Tracking
          & -- & -- & -- & -- & 15.277 & 0.672 & 0.446 \\
        $+$ Retrieval $+$ iPGO
          & $\times$ & $\times$ & -- & -- & 71.825 & 2.919 & 0.602 \\
        $+$ Geometric Check
          & $\checkmark$ & $\times$ & -- & -- & 7.534 & 0.605 & 0.424 \\
        $+$ GNC-TLS
          & $\checkmark$ & $\checkmark$ & -- & -- & 7.568 & 0.604 & 0.423 \\
        $+$ Dense Opt.
          & $\checkmark$ & $\checkmark$ & $\checkmark$ & Free
          & 12.962 & \textbf{0.289} & \textbf{0.302} \\
        \textbf{Full method}
          & $\checkmark$ & $\checkmark$ & $\checkmark$ & Fixed
          & \textbf{7.489} & \underline{0.296} & \underline{0.306} \\
        \bottomrule
      \end{tabular}%
    }%
  }

  \noindent
  \begin{minipage}[t]{0.54\textwidth}
    \vspace{0pt}
    \centering
    (a) System component ablation on KITTI\par
    \vspace{2pt}
    \usebox{\systemablationtablebox}
  \end{minipage}\hfill
  \begin{minipage}[t]{0.452\textwidth}
    \vspace{0pt}
    \centering
    (b) Robust loop selection with GNC-TLS\par
    \vspace{2pt}
    \includegraphics[
      width=\linewidth,
      height=\dimexpr\ht\systemablationtablebox+\dp\systemablationtablebox\relax,
      keepaspectratio
    ]{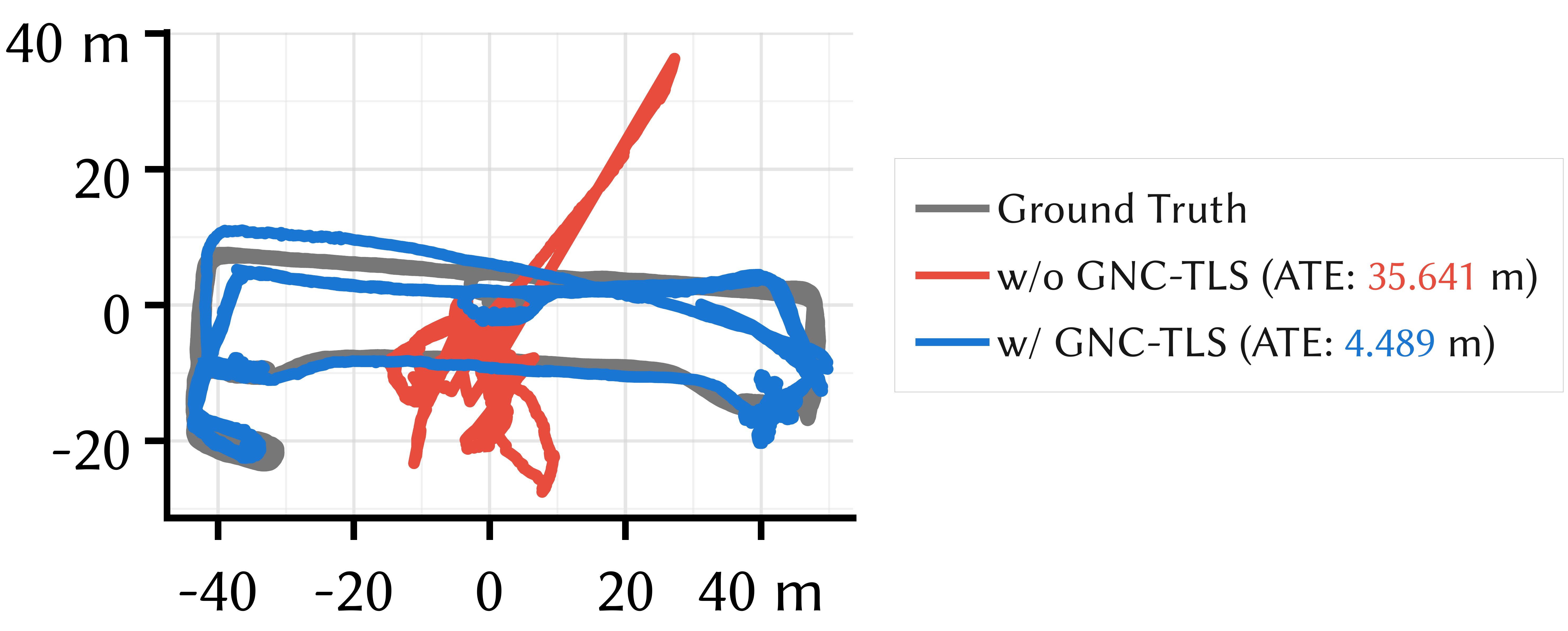}
  \end{minipage}

  \vspace{2pt}
  \noindent
  \begin{minipage}{\textwidth}
    \centering
    {\setlength{\tabcolsep}{1pt}%
    \begin{tabular}{@{}ccc@{}}
      \systemablationstage{Tracking}{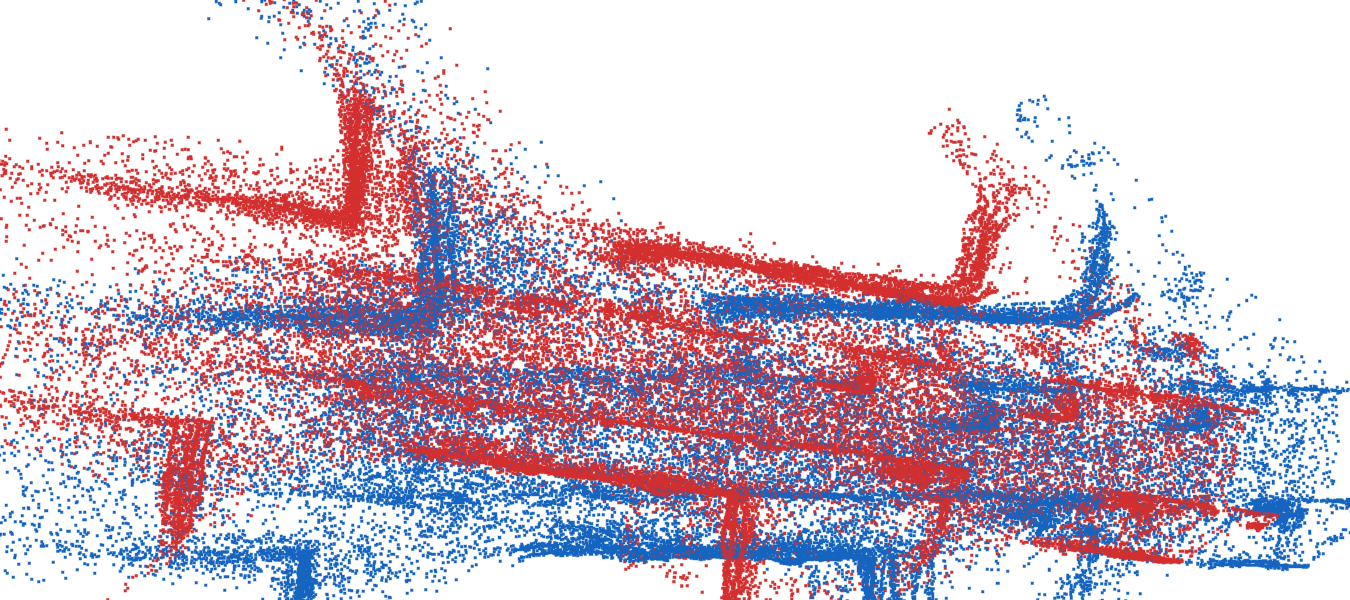} &
      \systemablationstage{iPGO}{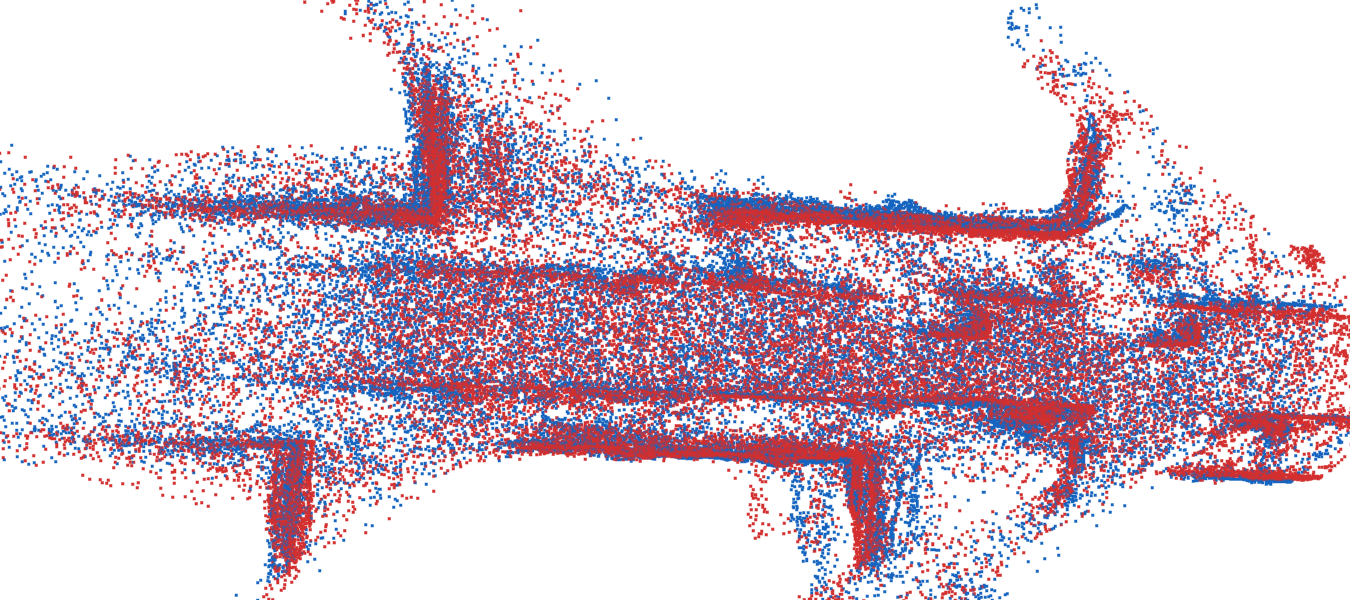} &
      \systemablationstage{Dense Opt.}{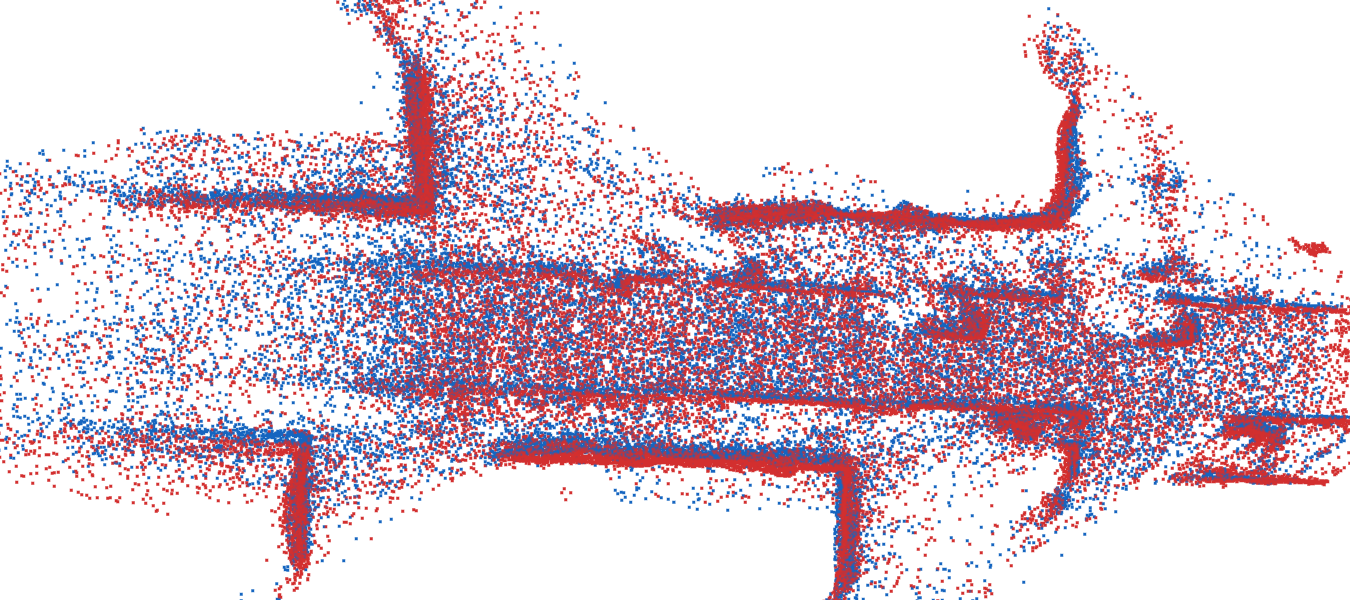}
    \end{tabular}}\par
    \vspace{1pt}
    (c) Loop-region geometry across system stages
  \end{minipage}

  \captionsetup{skip=6pt}
  \caption{\textbf{System Ablations.} \textbf{(a)} Mean results on KITTI sequences 00--10 show that Geometric Check filters false loop candidates, while fixing segment alignments to the solution from Incremental Pose Graph Optimization (iPGO) recovers low ATE and retains low RPE. \textbf{(b)} On VBR \emph{diag\_train0}, GNC-TLS avoids the trajectory failure observed without it, reducing ATE from \(35.641\,\mathrm{m}\) to \(4.489\,\mathrm{m}\). \textbf{(c)} Across Tracking, iPGO, and Dense Pose--Depth Optimization (Dense Opt.), the same loop-region point samples from the earlier (blue) and later (red) visits are overlaid in a common coordinate frame.}
  \label{fig:system-ablation}
\end{figure*}

\paragraph{Pairwise Geometry}
Geometric consistency verification is critical for filtering false loop
candidates before iPGO. As shown in Figure~\ref{fig:system-ablation}(a),
without this check, adding Loop Retrieval and iPGO increases the mean ATE from
\(15.277\,\mathrm{m}\) to \(71.825\,\mathrm{m}\); with Pairwise Geometry, it
drops to \(7.534\,\mathrm{m}\).

\paragraph{GNC-TLS}
Pairwise-verified loops can still conflict globally. Although GNC-TLS has
little effect on the KITTI mean ATE
\((7.534 \rightarrow 7.568\,\mathrm{m})\), it prevents the trajectory failure
in the VBR \emph{diag\_train0} stress case in
Figure~\ref{fig:system-ablation}(b). Its benefit therefore lies in robustness
to conflicting loop constraints rather than average accuracy.

\paragraph{Dense Geometric Refinement}
Free segment alignment substantially reduces translational and rotational RPE
from \(0.604\,\mathrm{m}/0.423^\circ\) to
\(0.289\,\mathrm{m}/0.302^\circ\), but degrades the mean ATE from
\(7.568\,\mathrm{m}\) to \(12.962\,\mathrm{m}\), showing that better local
geometry can come at the expense of global consistency. Fixing the
iPGO-conditioned alignments restores the mean ATE to \(7.489\,\mathrm{m}\)
while retaining low RPE at \(0.296\,\mathrm{m}/0.306^\circ\). Dense refinement
is applied only at sequence end; online global correction is provided by
iPGO.

\subsection{Analysis of the Loop Retrieval Descriptor}
\label{sec:vpr-evaluation}

\begin{figure*}[t]
  \centering
  \includegraphics[
    width=5.5in,
    trim=0 86bp 0 8bp,
    clip
  ]{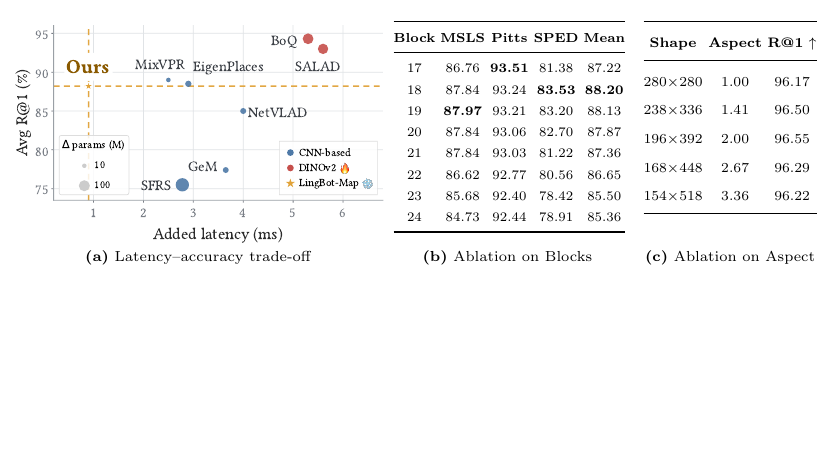}
  \captionsetup{skip=3pt}
  \caption{\textbf{Visual place retrieval with LingBot-Map's Frame Encoder.}
  All three panels report Recall@1 (R@1 \(\uparrow\)).
  \textbf{(a)} By reusing cached Frame Encoder features, our full SLAM system
  achieves VPR accuracy comparable to strong CNN-based baselines with only a
  1.8M-parameter head and \(0.94\,\mathrm{ms}\) of additional descriptor
  latency for \(322\times322\) inputs. \textbf{(b)} Block 18 is the best
  retrieval tap among the last eight Frame Encoder blocks. \textbf{(c)} The
  square-trained head remains stable across aspect ratios, allowing retrieval
  to consume cached features directly at the runtime input
  shape.}
  \label{fig:vpr-analysis}
\end{figure*}

Our retrieval descriptor reuses cached token features captured during
LingBot-Map's Frame Encoder pass. A 1.8M-parameter SALAD aggregation head
converts these tokens into a global descriptor~\citep{izquierdo2024optimal}.
For \(322\times322\) input images, this conversion takes only
\(0.94\,\mathrm{ms}\). We train only this head on
GSV-Cities~\citep{ali2022gsv} for 10 epochs at \(224\times224\) resolution
while keeping the Frame Encoder frozen. Training takes approximately 2 hours
on one NVIDIA RTX 5090D GPU with mixed precision. The resulting descriptor
achieves a mean Recall@1 (R@1)
of 88.2 across MSLS-val~\citep{warburg2020mapillary},
Pitts250k-test~\citep{torii2013visual}, and SPED~\citep{chen2017deep}, as shown
in Figure~\ref{fig:vpr-analysis}(a). This result is comparable to strong
CNN-based VPR baselines such as EigenPlaces~\citep{berton2023eigenplaces}
(88.5) and MixVPR~\citep{ali2023mixvpr} (89.0). Although it trails standalone
DINOv2 models specialized for VPR, including
SALAD~\citep{izquierdo2024optimal} and BoQ~\citep{ali2024boq}, the descriptor
remains effective as the loop-candidate proposal stage of our full system.
Pairwise Geometry verifies retrieved pairs before they enter the pose graph,
and the complete-system results are reported in
Tables~\ref{tab:kitti-results}, \ref{tab:vbr-results}, and
\ref{tab:oxford-spires-results}. 

Among the tested late Frame Encoder blocks, intermediate features outperform
the final-block features for retrieval. Block 18 achieves the highest mean R@1
of 88.20, closely followed by Block 19 at 88.13, whereas the final Block 24
reaches 85.36 (Figure~\ref{fig:vpr-analysis}(b)). This trend is consistent with
prior observations that DINO-based VPR can peak before the final transformer
layer~\citep{keetha2023anyloc}. Finally, we test whether the square-trained head
transfers to the native wide input used by the SLAM system. On
KITTI~\citep{geiger2012we}, R@1 varies by only 0.38 points across five
approximately token-matched input shapes and reaches 96.22 at the
\(154\times518\) runtime shape (Figure~\ref{fig:vpr-analysis}(c)). These results
support applying the retrieval head directly to cached intermediate DINO
tokens at the runtime input shape.

\section{Conclusions}
\label{sec:conclusions}

We presented Map2SLAM, which builds SLAM on a frozen LingBot-Map by
scheduling its inference and keeping its predictions as revisable state.
The front-end updates this state through Causal Continuation 
for stable online tracking, while the asynchronous back-end uses Bidirectional Prefix 
with newly available context to re-evaluate loop constraints for iPGO and to re-observe segments for Dense Geometric Refinement after the sequence. This design preserves local tracking stability while allowing global trajectory 
errors to be corrected over time. Experiments 
on KITTI, VBR, and Oxford Spires show that Map2SLAM reduces mean ATE by more than 40\% compared with the second-best uncalibrated method, while maintaining a tracking speed above 28 FPS on an NVIDIA RTX 5090D GPU.

\subparagraph*{AI use statement}

Generative AI tools were used to assist with manuscript drafting and language editing, as well as software development, including code generation, debugging, testing, and execution of experiments under instructions provided by the authors. The research ideas, methodology, experimental design, and interpretation of results were developed by the authors. All AI-assisted text, code, and experimental outputs were reviewed and verified by the authors. The authors take full responsibility for the final content and results of this work.

\bibliography{references}
\bibliographystyle{manuscript}

\appendix
\clearpage

\section{Additional Benchmark Results}
\label{sec:additional-benchmark-results}

Table~\ref{tab:oxford-spires-results} reports the per-sequence results on
Oxford Spires. All methods process each sequence at its full length, without
temporal cropping. We compute ATE after a single global \(\mathrm{Sim}(3)\)
alignment using all predicted poses with valid ground-truth correspondences.

\begin{table}[!htbp]
  \centering
  \captionsetup{skip=2pt}
  \caption{Absolute trajectory error (ATE, m) on the Oxford Spires
  Dataset~\citep{tao2026oxford}. KC, OQ, CC, and BL denote Keble College,
  Observatory Quarter, Christ Church, and Bodleian Library, respectively.
  ``--'' denotes tracking failure. LC, Dense Geo., and Frame-wise denote loop
  closure, dense geometry, and per-frame tracking, respectively.
  LingBot-Map(S/W) denote the streaming/window modes. $^\dagger$ VGGT-SLAM 2.0
  uses native keyframes only and is excluded from ranking. The best and
  second-best results in each section are shown in bold and underlined,
  respectively.}
  \label{tab:oxford-spires-results}
  \scriptsize
  \setlength{\tabcolsep}{2.0pt}
  \renewcommand{\arraystretch}{1.18}
  \newcommand{\oxseqmeta}[1]{{\color{gray}\itshape\fontsize{4.2}{4.6}\selectfont #1}}
  \newcommand{\oxseqhead}[3]{%
    \shortstack{\rule{0pt}{2.4ex}\textbf{#1}\\[0.45ex]
    \oxseqmeta{#2}\\[0.10ex]
    \oxseqmeta{#3}\rule[-0.9ex]{0pt}{0.9ex}}}
  \resizebox{\linewidth}{!}{%
  \begin{tabular}{@{}l|c|cc|cccccccccc|c@{}}
    \toprule
    \raisebox{1.7ex}{\textbf{Method}} & \raisebox{1.7ex}{\textbf{LC}} &
    \raisebox{0.7ex}{\shortstack{\textbf{Dense}\\\textbf{Geo.}}} &
    \raisebox{0.7ex}{\shortstack{\textbf{Frame-}\\\textbf{wise}}} &
    \oxseqhead{KC-02}{5969\,f}{0.29\,km} &
    \oxseqhead{KC-03}{5682\,f}{0.28\,km} &
    \oxseqhead{KC-04}{13518\,f}{0.78\,km} &
    \oxseqhead{KC-05}{11544\,f}{0.71\,km} &
    \oxseqhead{OQ-01}{5746\,f}{0.40\,km} &
    \oxseqhead{OQ-02}{5469\,f}{0.39\,km} &
    \oxseqhead{CC-02}{11775\,f}{0.64\,km} &
    \oxseqhead{CC-03}{6204\,f}{0.27\,km} &
    \oxseqhead{CC-05}{3961\,f}{0.81\,km} &
    \oxseqhead{BL-02}{4966\,f}{0.69\,km} &
    \raisebox{1.7ex}{\textbf{Avg. $\downarrow$}} \\
    \midrule
    \multicolumn{15}{@{}l}{\emph{Calibrated}} \\
    DROID-SLAM~\citep{teed2021droid} & $\checkmark$ & $\checkmark$ & $\checkmark$
      & \underline{7.89} & \underline{31.50} & \underline{3.93} & \underline{36.44} & \textbf{0.43} & \textbf{0.29} & \underline{4.09} & \underline{0.86} & \textbf{8.96} & \textbf{1.20} & \underline{9.56} \\
    DPV-SLAM++~\citep{lipson2024deep} & $\checkmark$ & & $\checkmark$
      & \textbf{2.62} & \textbf{0.41} & \textbf{1.29} & \textbf{0.53} & \underline{0.74} & \underline{0.50} & \textbf{1.77} & \textbf{0.08} & \underline{11.93} & \underline{2.91} & \textbf{2.28} \\
    \midrule
    \multicolumn{15}{@{}l}{\emph{Uncalibrated}} \\
    DROID-Calib~\citep{hagemann2023deep} & & $\checkmark$ & $\checkmark$
      & 11.89 & 4.82 & \textbf{1.83} & 37.02 & \textbf{1.01} & 3.24 & 9.92 & \underline{0.63} & \textbf{2.63} & \textbf{1.36} & 7.43 \\
    VGGT-Long~\citep{deng2026vggt-long} & $\checkmark$ & $\checkmark$ &
      & \underline{1.76} & 14.74 & 32.53 & \textbf{4.59} & 3.70 & 5.66 & 11.65 & 4.00 & 13.51 & 16.16 & 10.83 \\
    SwiftVGGT~\citep{lee2026swiftvggt} & $\checkmark$ & $\checkmark$ &
      & 7.71 & 22.69 & 24.53 & 21.13 & 7.01 & 6.18 & 25.18 & 0.89 & 10.77 & 29.74 & 15.58 \\
    VGGT-SLAM 2.0$^\dagger$~\citep{maggio2026vggt-slam2} & $\checkmark$ & $\checkmark$ &
      & 4.63 & 9.32 & 4.75 & 5.98 & 1.84 & 1.65 & 1.99 & 0.93 & 36.25 & 8.32 & 7.57 \\
    LoGeR~\citep{zhang2026loger} & & $\checkmark$ &
      & 4.23 & 3.77 & 6.87 & 7.19 & 5.46 & 6.05 & \underline{6.68} & 2.81 & 11.75 & 14.01 & 6.88 \\
    LingBot-Map(S)~\citep{chen2026geometric} & & $\checkmark$ & $\checkmark$
      & 3.72 & \underline{3.02} & 2.73 & 6.90 & 3.26 & 2.29 & 12.56 & 2.79 & 33.76 & 10.38 & 8.14 \\
    LingBot-Map(W)~\citep{chen2026geometric} & & $\checkmark$ & $\checkmark$
      & 5.40 & 8.11 & 36.67 & 29.71 & 7.76 & 6.95 & 15.36 & 4.98 & 14.45 & 29.92 & 15.93 \\
    HorizonStream~\citep{cheng2026horizonstream} & $\checkmark$ & $\checkmark$ & $\checkmark$
      & 6.82 & 8.84 & 21.19 & 21.94 & 5.99 & \textbf{1.54} & 30.19 & 5.84 & 26.29 & 27.97 & 15.66 \\
    ABot-Recon~\citep{han2026revisiting} & $\checkmark$ & $\checkmark$ & $\checkmark$
      & 2.36 & 4.25 & 5.12 & \underline{4.65} & 3.71 & 2.86 & 7.94 & 1.92 & 9.37 & 11.41 & \underline{5.36} \\
    \textbf{Ours} & $\checkmark$ & $\checkmark$ & $\checkmark$
      & \textbf{1.53} & \textbf{1.89} & \underline{2.39} & 5.73 & \underline{2.05} & \underline{1.78} & \textbf{2.68} & \textbf{0.52} & \underline{6.32} & \underline{5.82} & \textbf{3.07} \\
    \bottomrule
  \end{tabular}}
\end{table}

\section{SLAM System Profiling}

\begin{figure}[htbp]
  \centering
  \includegraphics[width=\linewidth]{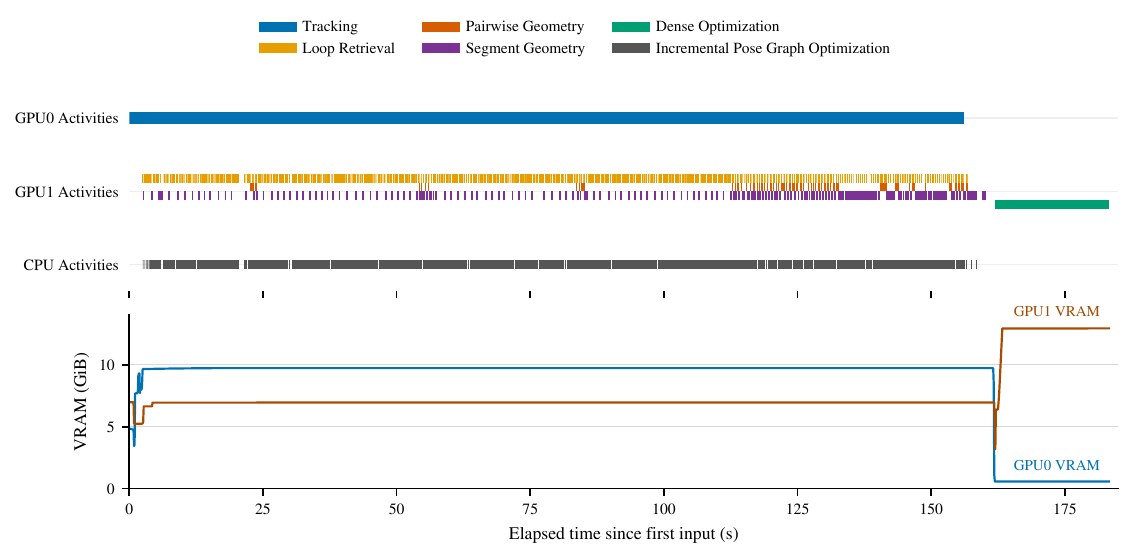}
  \captionsetup{skip=0pt}
  \caption{\textbf{System execution and memory profile for a complete KITTI 00
  run on two NVIDIA RTX 5090D GPUs.} During the online stage, GPU 0 performs
  Tracking, GPU 1 interleaves Loop Retrieval, Pairwise Geometry, and Segment
  Geometry, and the CPU executes Incremental PGO. GPU 1 performs terminal Dense
  Optimization after the online stage. The lower plot reports per-GPU VRAM
  usage throughout the run.}
  \label{fig:system-profile}
\end{figure}

We profile the complete KITTI Sequence~00 on two NVIDIA RTX 5090D GPUs.
GPU0 and GPU1 are dedicated to the front end and back end, respectively.
GPU0 executes Tracking, while GPU1 executes Loop Retrieval, Pairwise Geometry,
Segment Geometry, and terminal Dense Optimization.
Both Pairwise Geometry and Segment Geometry use fresh-context bidirectional
prefix inference over selected historical frames to obtain repeated geometric
observations.
On the CPU, we record only Incremental Pose Graph Optimization (iPGO).
The activity bars show the measured execution spans of these components, while
the memory curves report GPU memory usage sampled every \(100~\mathrm{ms}\).

The profile shows the following:
\begin{enumerate}
    \item \textbf{Smooth front-end tracking.}
    GPU0 remains almost continuously occupied by Tracking and processes
    KITTI Sequence~00 at \(29.1~\mathrm{FPS}\), indicating steady front-end
    execution.

    \item \textbf{Asynchronous front-end/back-end execution.}
    Tracking on GPU0 overlaps with loop processing and Segment Geometry on
    GPU1, as well as iPGO on the CPU.
    The repeated iPGO activity shows that the trajectory is updated online
    while new frames are being tracked.

    \item \textbf{Effective loop-candidate filtering.}
    From \(4{,}541\) input frames, the system selects \(2{,}208\) keyframes
    and computes a SALAD descriptor for each, but issues only \(429\)
    retrieval queries.
    These queries produce \(2{,}141\) candidate pairs, of which only \(175\)
    (\(8.2\%\)) enter Pairwise Geometry.
    Among them, \(59\) pass verification, corresponding to \(33.7\%\) of the
    geometrically verified candidates and \(2.8\%\) of all retrieved pairs.
    The retrieval and candidate-pool filters therefore restrict expensive
    Pairwise Geometry to a small subset of candidate pairs.

    \item \textbf{Opportunistic Segment Geometry.}
    Loop Retrieval and Pairwise Geometry occupy only short intervals on GPU1,
    leaving substantial idle capacity for Segment Geometry.
    By running Segment Geometry concurrently with tracking, most segment
    observations are completed online, and the final Segment Geometry
    execution finishes only approximately \(4.15~\mathrm{s}\) after tracking
    ends.

    \item \textbf{Dense-optimization memory bottleneck.}
    During online operation, GPU0 and GPU1 peak at \(9.73~\mathrm{GiB}\) and
    \(6.98~\mathrm{GiB}\), respectively.
    Terminal Dense Optimization raises the GPU1 peak to
    \(12.94~\mathrm{GiB}\).
    Reducing the depth-map resolution may therefore lower the peak memory
    requirement.
\end{enumerate}

\begin{figure}[htbp]
  \centering
  \includegraphics[width=\linewidth]{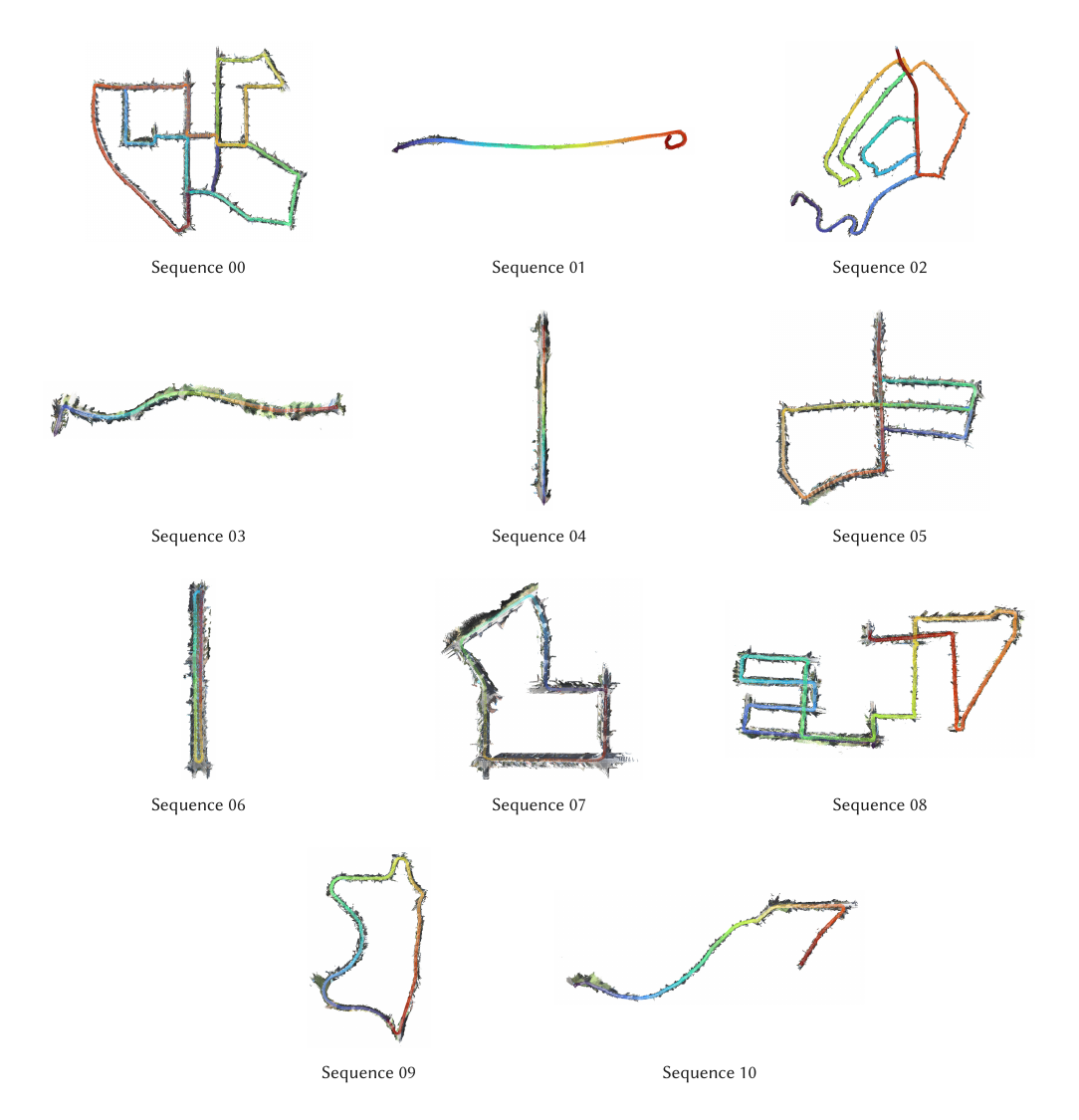}
  \captionsetup{skip=0pt}
  \captionsetup{skip=0pt}
  \caption{\textbf{Qualitative point-cloud results on KITTI.}Top-down views of the point clouds produced by our method for Sequences 00--10.
  }
  \label{fig:kitti_viz}
\end{figure}

\section{Detailed VPR Evaluation}
\label{sec:vpr-evaluation-supp}

Table~\ref{tab:vpr-accuracy-supp} extends the main-paper analysis with the
complete R@1/R@5/R@10 comparison on
MSLS-val~\citep{warburg2020mapillary},
Pitts250k-test~\citep{torii2013visual}, and SPED~\citep{chen2017deep}.
We measure every method within the
same evaluation framework, using
the same benchmark splits, ground-truth annotations, and FAISS L2 retrieval,
while retaining each method's standard evaluation resolution. Latency is
measured with batch size 1 and FP16 inference on a single NVIDIA RTX 5090D
GPU using CUDA events after at least 20 warm-up iterations. For standalone
baselines, the reported latency covers a full descriptor-model forward at each
method's standard evaluation resolution. For our method, it covers only the
1.8M-parameter aggregation-head forward on precomputed tokens from
\(322\times322\) images. GeM and NetVLAD are
our GSV-Cities reproductions because no standard checkpoints are publicly
available; the remaining rows use the corresponding released weights.

We train only the 1.8M-parameter aggregation head while keeping the Frame
Encoder frozen. Training runs for 10 epochs on
GSV-Cities~\citep{ali2022gsv} at \(224\times224\) resolution using 16-bit mixed
precision on one NVIDIA RTX 5090D GPU and takes approximately two hours. We
use AdamW with an initial learning rate of \(10^{-3}\) and a batch of 128
places with four images per place. The learning rate decays linearly to
\(2\times10^{-4}\) over training. We validate after each epoch on Pitts30k-val
and select the checkpoint with the highest R@1.

On MSLS-val, Pitts250k-test, and SPED, our descriptor reaches R@1 scores of
87.8, 93.2, and 83.5, respectively. It outperforms
SFRS~\citep{ge2020self}, GeM~\citep{radenovic2018fine},
Conv-AP~\citep{ali2022gsv}, and CosPlace~\citep{berton2022rethinking} on all
three datasets, while remaining within 1.2 points of MixVPR on each dataset.

\begin{table}[htbp]
  \centering
  \captionsetup{skip=2pt}
  \caption{\textbf{Detailed visual place retrieval results.}
  Each dataset reports R@1/R@5/R@10 (\%); higher is better.
  \(\Delta\)Params and \(\Delta\)Latency denote the additional model
  parameters and inference latency incurred by adding each VPR method to our
  SLAM system. Standalone baselines add their full descriptor model, whereas
  Ours adds only the aggregation head on cached Frame Encoder tokens.}
  \label{tab:vpr-accuracy-supp}
  \scriptsize
  \setlength{\tabcolsep}{2.2pt}
  \renewcommand{\arraystretch}{1.16}
  \resizebox{\linewidth}{!}{%
  \begin{tabular}{@{}lrrrccc@{}}
    \toprule
    \textbf{Method} & \textbf{Dim.} & \textbf{\(\Delta\)Params (M)} &
    \textbf{\(\Delta\)Latency (ms)} & \textbf{MSLS-val} &
    \textbf{Pitts250k-test} & \textbf{SPED} \\
    \midrule
    SFRS~\citep{ge2020self} & 4096 & 149.0 & 2.78 & 64.2/74.6/77.8 & 89.5/95.3/96.5 & 73.0/85.7/87.8 \\
    GeM~\citep{radenovic2018fine} & 2048 & 23.5 & 3.65 & 78.9/87.6/90.1 & 86.5/94.6/96.4 & 66.9/79.4/85.2 \\
    NetVLAD~\citep{arandjelovic2016netvlad} & 32768 & 24.6 & 4.03 & 83.1/90.7/92.7 & 92.6/97.5/98.3 & 79.2/89.3/91.9 \\
    Conv-AP~\citep{ali2022gsv} & 4096 & 25.6 & 2.92 & 83.0/90.1/92.6 & 92.6/97.7/98.6 & 81.1/89.6/93.3 \\
    CosPlace~\citep{berton2022rethinking} & 2048 & 27.7 & 3.08 & 87.6/92.7/94.7 & 92.4/97.5/98.4 & 79.6/89.0/92.6 \\
    MixVPR~\citep{ali2023mixvpr} & 4096 & 10.9 & 2.48 & 88.1/93.2/94.5 & 94.3/98.2/98.9 & 84.7/92.3/94.4 \\
    EigenPlaces~\citep{berton2023eigenplaces} & 2048 & 27.7 & 2.90 & 89.1/93.1/95.3 & 94.5/98.1/98.8 & 81.9/92.3/95.1 \\
    \midrule
    SALAD~\citep{izquierdo2024optimal} & 8448 & 88.0 & 5.57 & 92.0/96.2/96.9 & 95.0/98.5/99.2 & 91.9/96.1/96.7 \\
    CricaVPR~\citep{lu2024cricavpr} & 10752 & 106.8 & 13.55 & 89.5/95.4/96.2 & 95.6/98.5/99.2 & 89.6/94.6/95.4 \\
    BoQ (DINOv2)~\citep{ali2024boq} & 12288 & 95.2 & 5.34 & 93.8/96.8/97.0 & 96.6/99.1/99.5 & 92.6/95.9/96.9 \\
    MegaLoc~\citep{berton2025megaloc} & 8448 & 228.6 & 5.93 & 93.4/97.0/97.8 & 96.4/98.9/99.4 & 90.4/95.2/95.9 \\
    \midrule
    Ours & 8448 & 1.8 & 0.94 &
    87.8/93.8/94.6 & 93.2/97.6/98.6 &
    83.5/92.1/94.6 \\
    \bottomrule
  \end{tabular}}
\end{table}

\clearpage
\begin{figure*}[p]
  \centering
  \begingroup
  \newcommand{\supptrajectorypanel}[1]{%
    \includegraphics[width=0.316\linewidth]{#1}%
  }
  \newcommand{\suppscenelabel}[1]{%
    \begin{minipage}[c]{0.07\textwidth}
      \centering
      \rotatebox[origin=c]{90}{\small\bfseries #1}
    \end{minipage}%
  }
  \newcommand{\suppsceneblock}[6]{%
    \begin{minipage}[c]{0.73\textwidth}
      \centering
      \setlength{\tabcolsep}{0pt}
      \begin{tabular}{@{}ccc@{}}
        \supptrajectorypanel{#1} &
        \supptrajectorypanel{#2} &
        \supptrajectorypanel{#3} \\[-3pt]
        {\scriptsize (a) LoGeR} &
        {\scriptsize (b) HorizonStream + LC} &
        {\scriptsize (c) ABot-Recon + LC} \\[-1pt]
        \supptrajectorypanel{#4} &
        \supptrajectorypanel{#5} &
        \supptrajectorypanel{#6} \\[-3pt]
        {\scriptsize (d) LingBot-Map (S)} &
        {\scriptsize (e) LingBot-Map (W)} &
        {\scriptsize (f) Ours}
      \end{tabular}
    \end{minipage}%
  }
  \noindent\makebox[\dimexpr\linewidth-2pt\relax][c]{%
  \begin{tabular}{@{}c@{\hspace{6pt}}c@{}}
    \suppscenelabel{KITTI 00} &
    \suppsceneblock
      {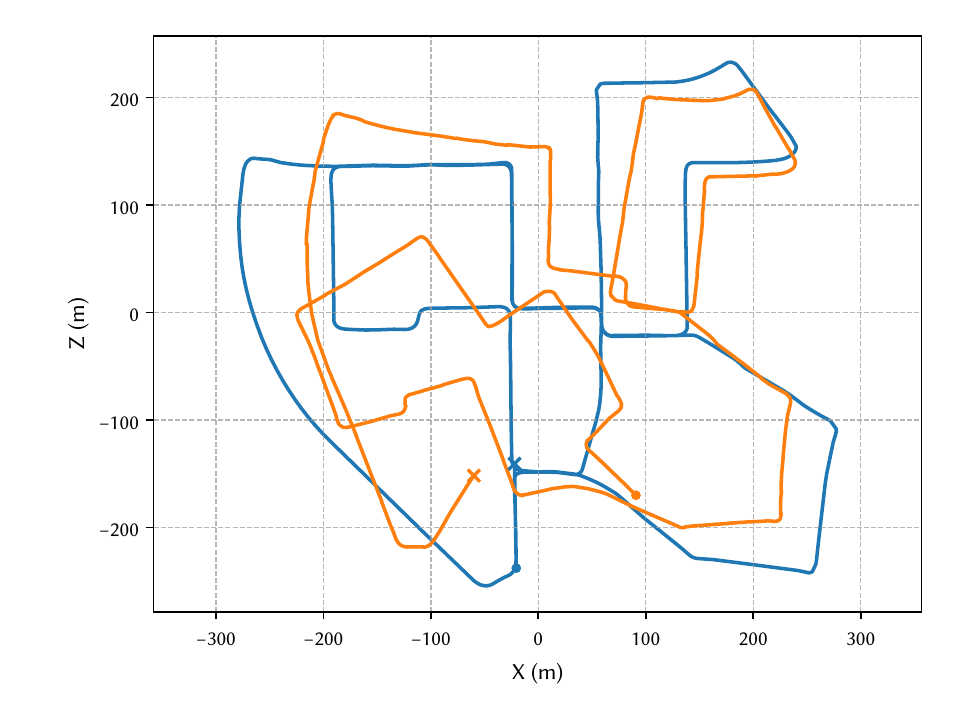}
      {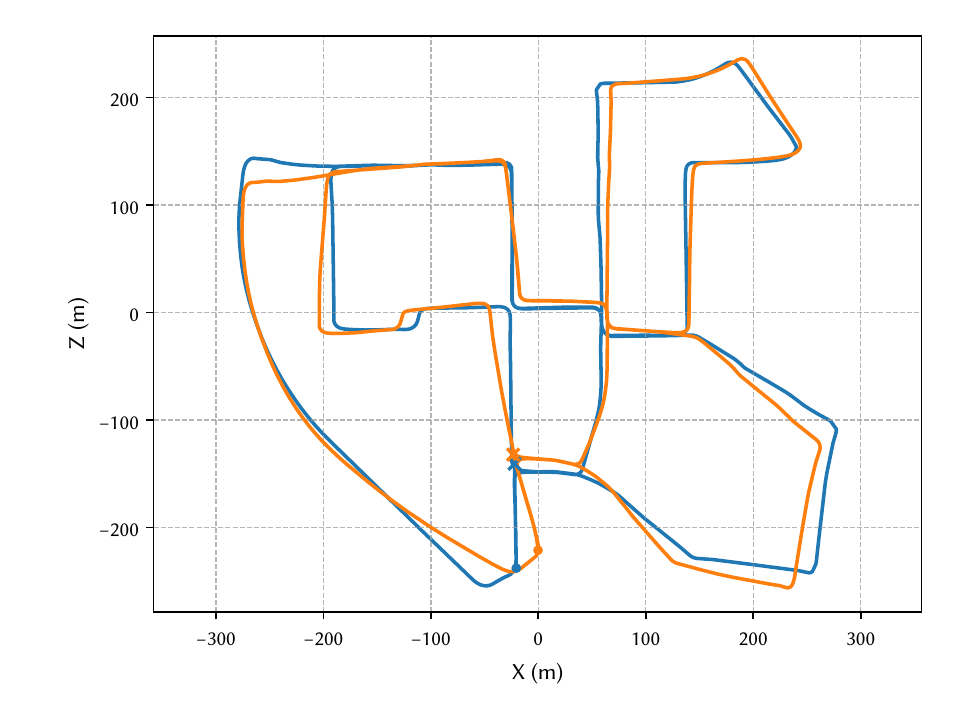}
      {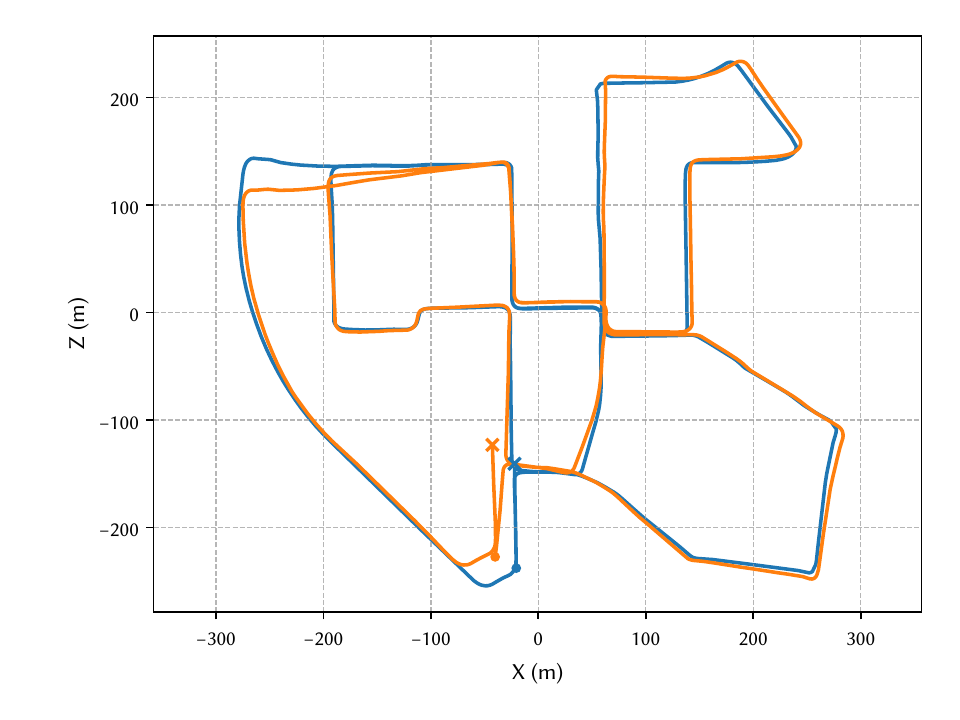}
      {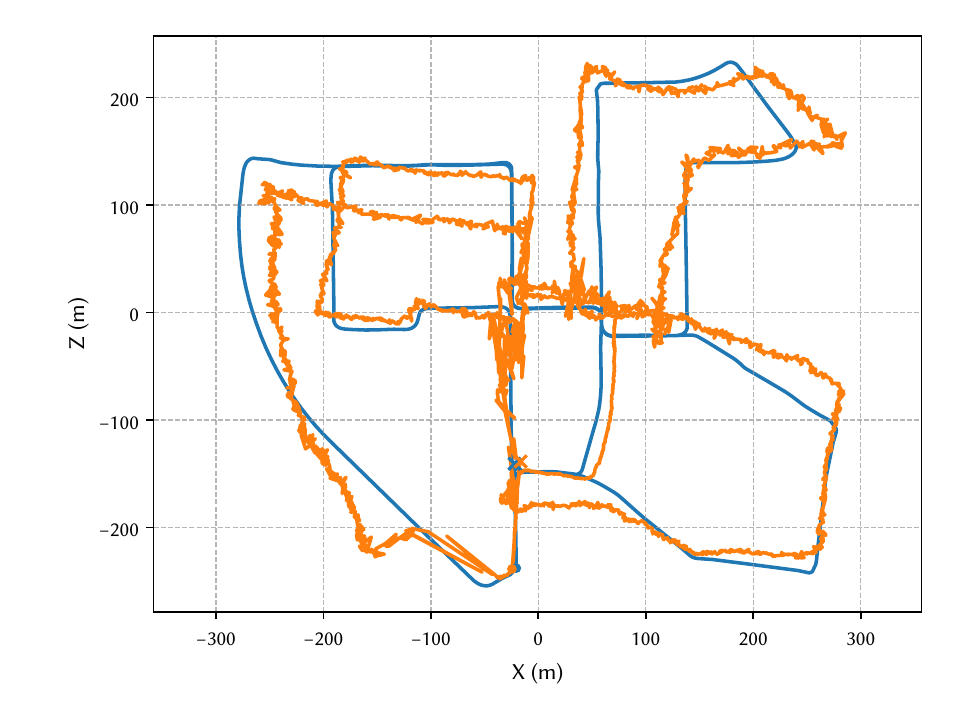}
      {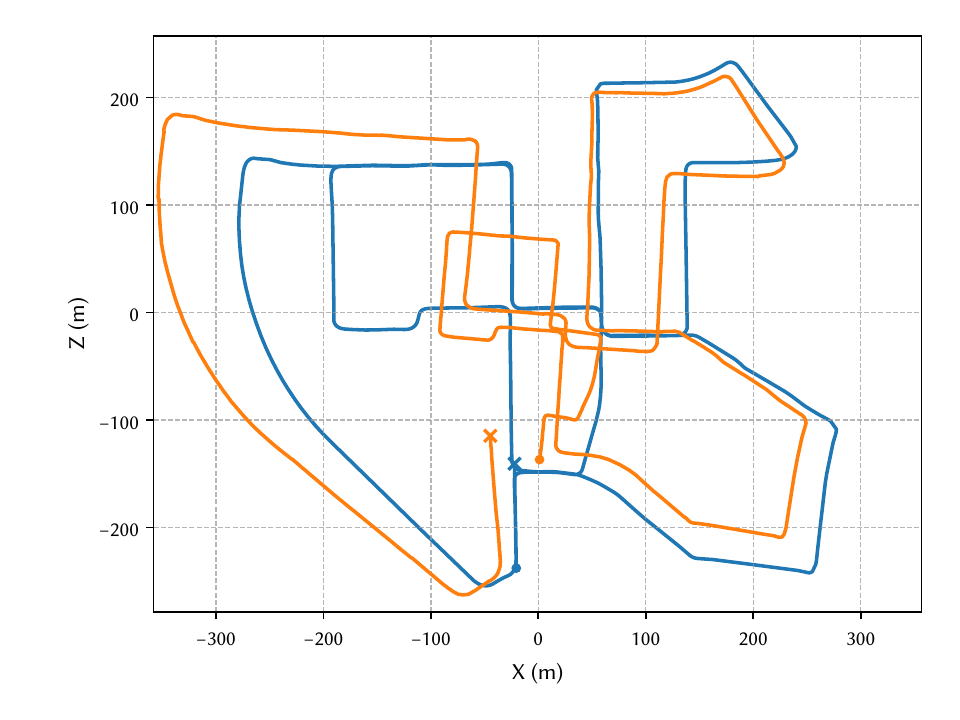}
      {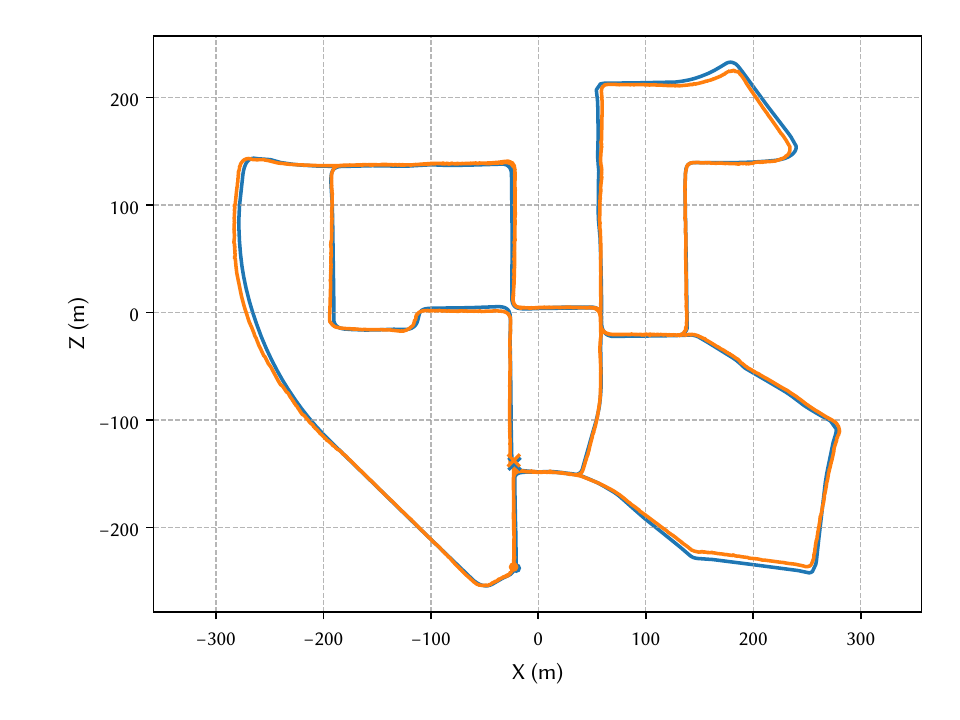} \\[3pt]
    \suppscenelabel{KITTI 02} &
    \suppsceneblock
      {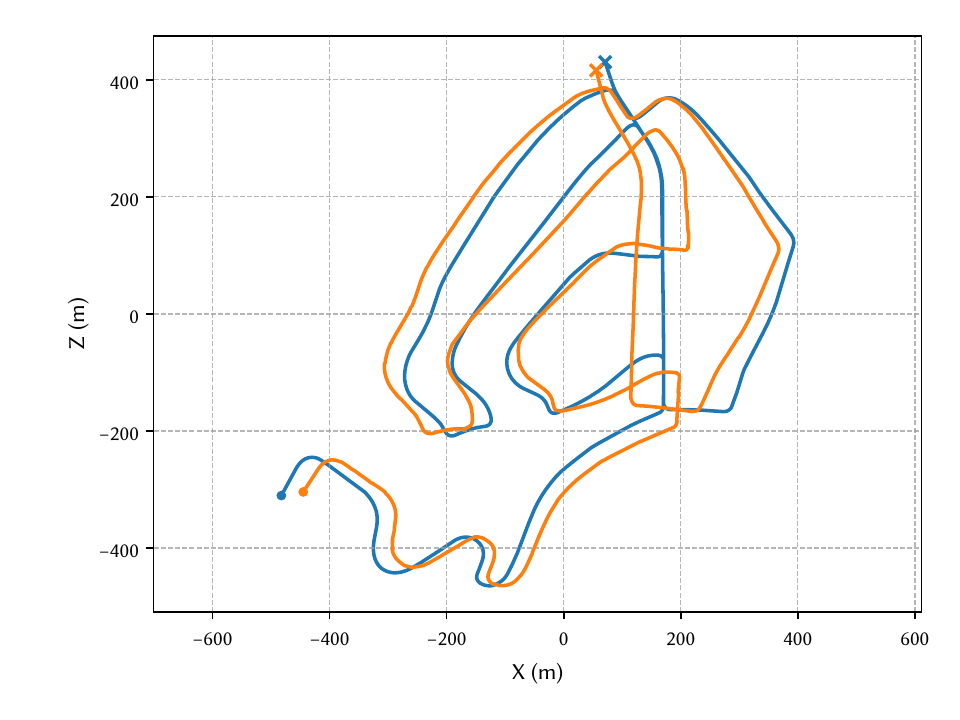}
      {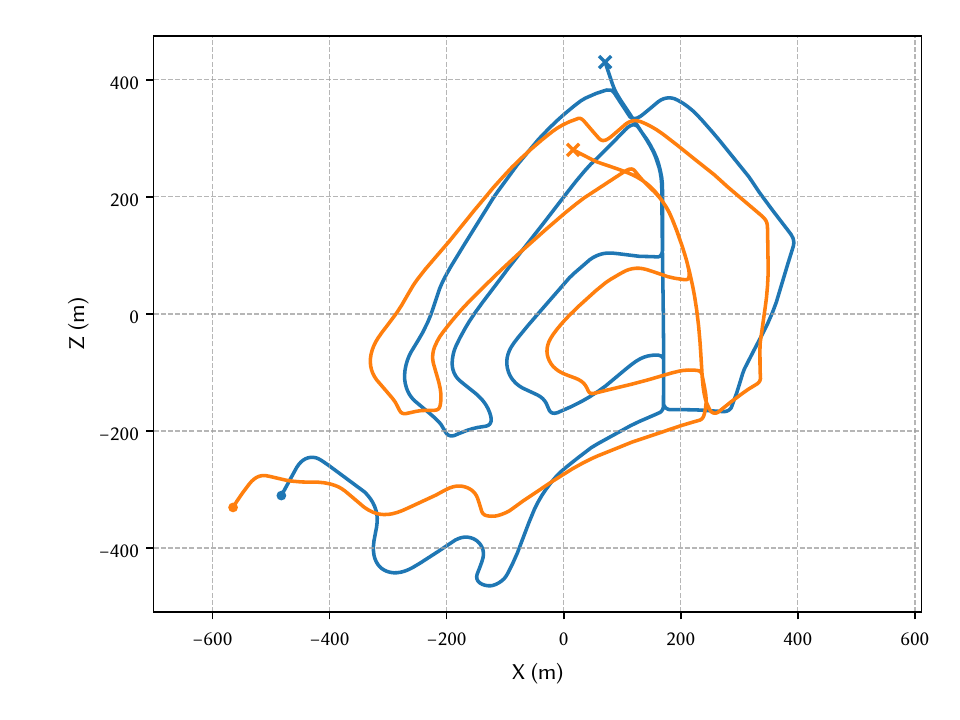}
      {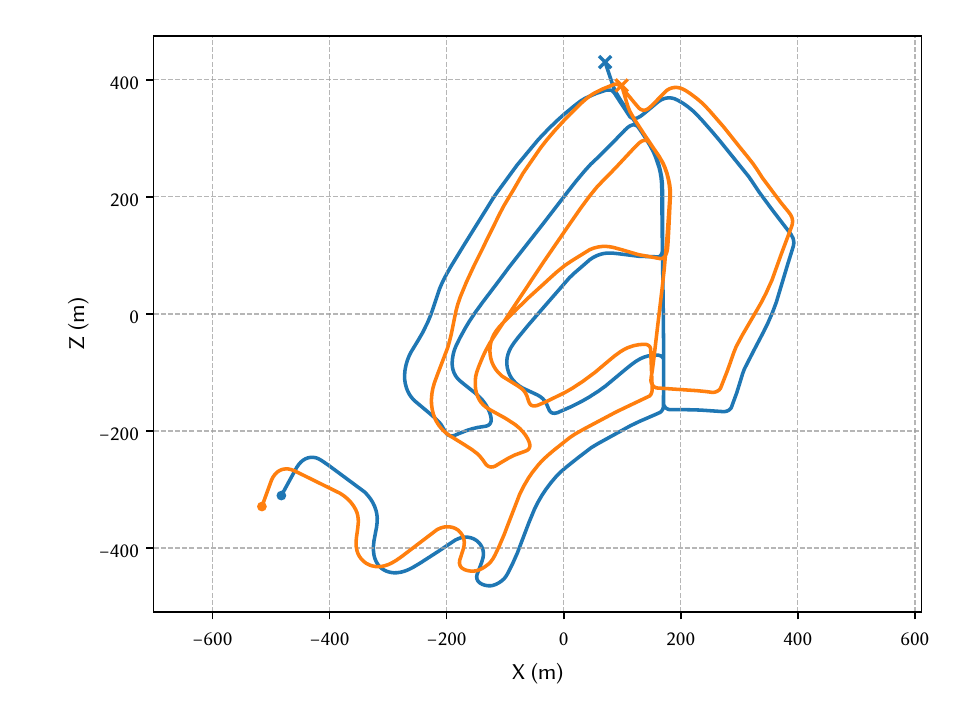}
      {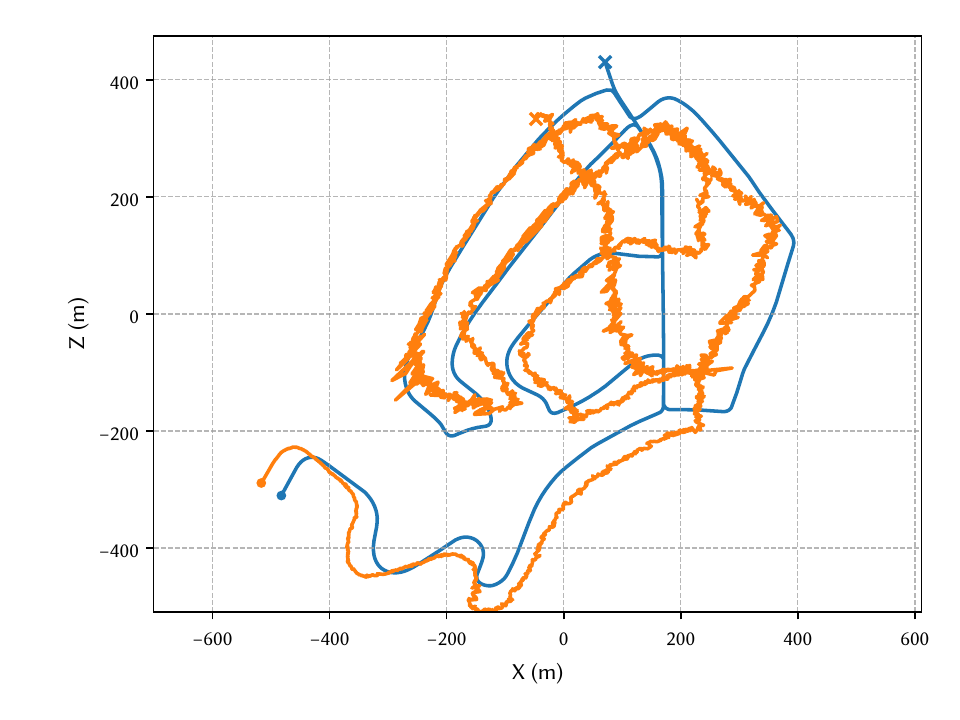}
      {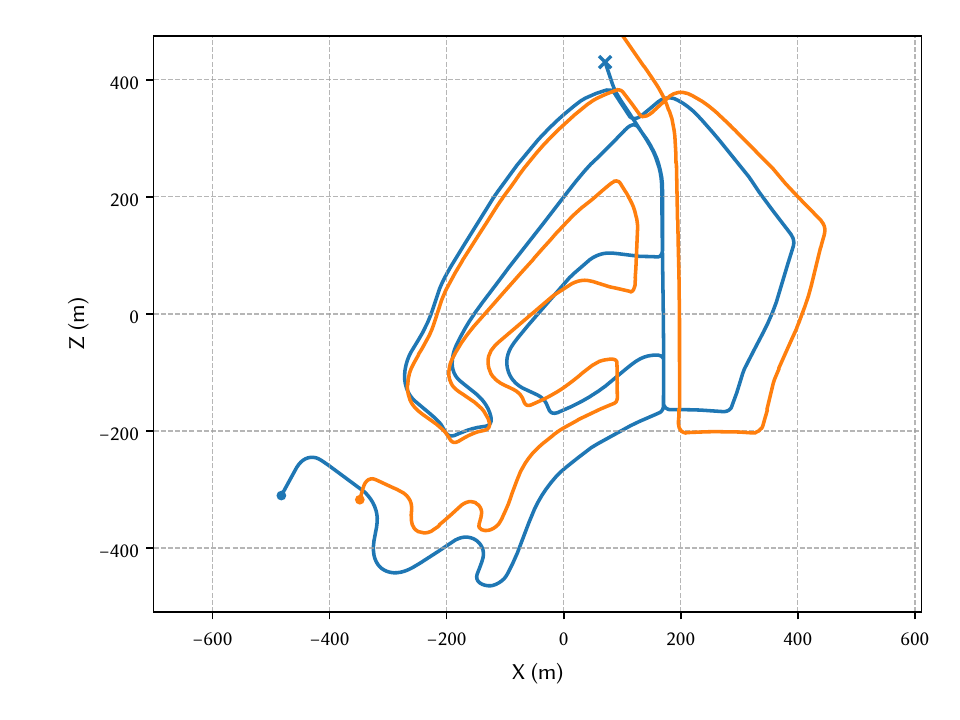}
      {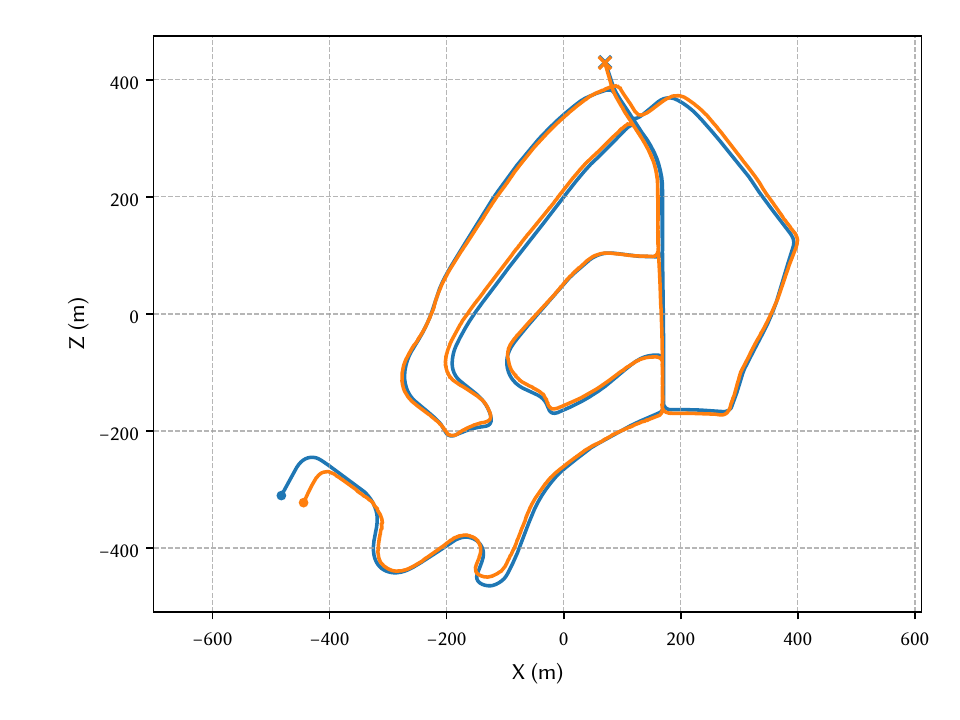} \\[3pt]
    \suppscenelabel{\shortstack{Oxford Spires\\\emph{Christ Church 05}}} &
    \suppsceneblock
      {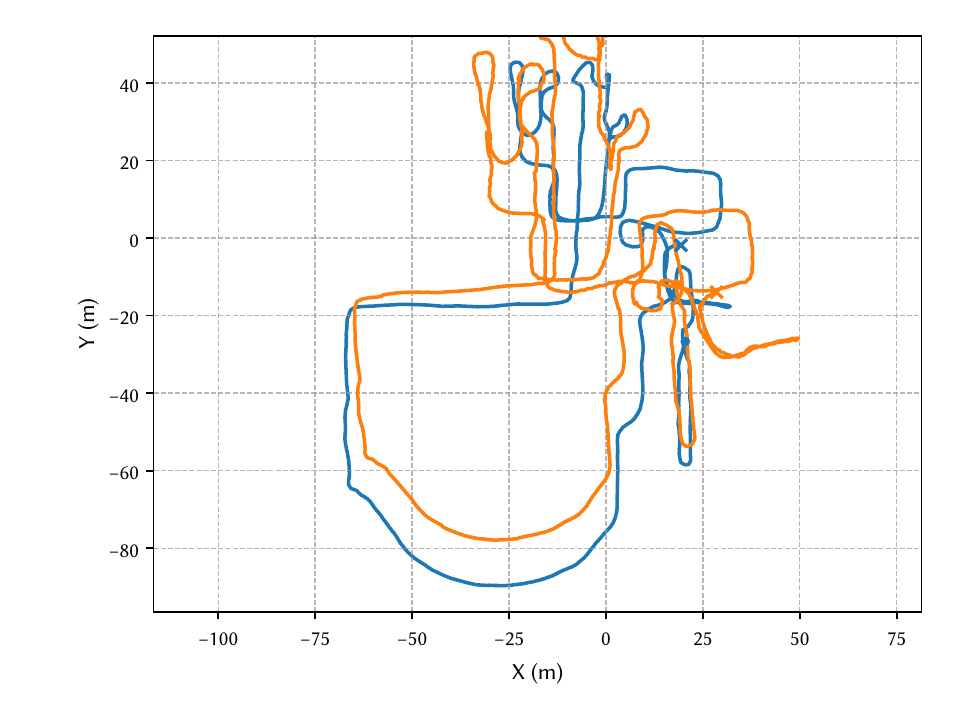}
      {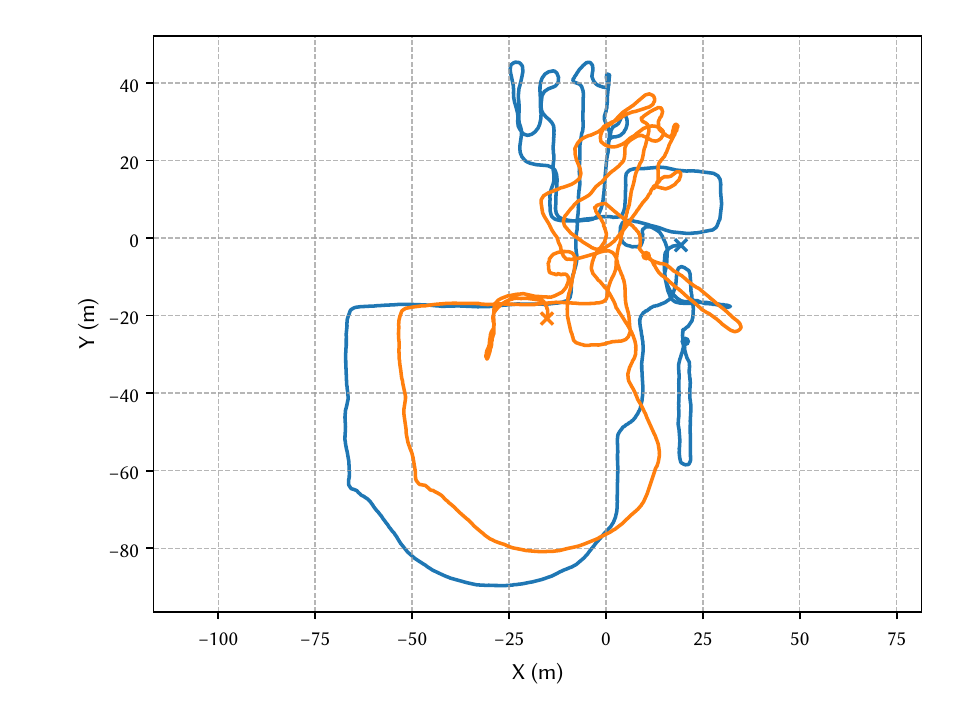}
      {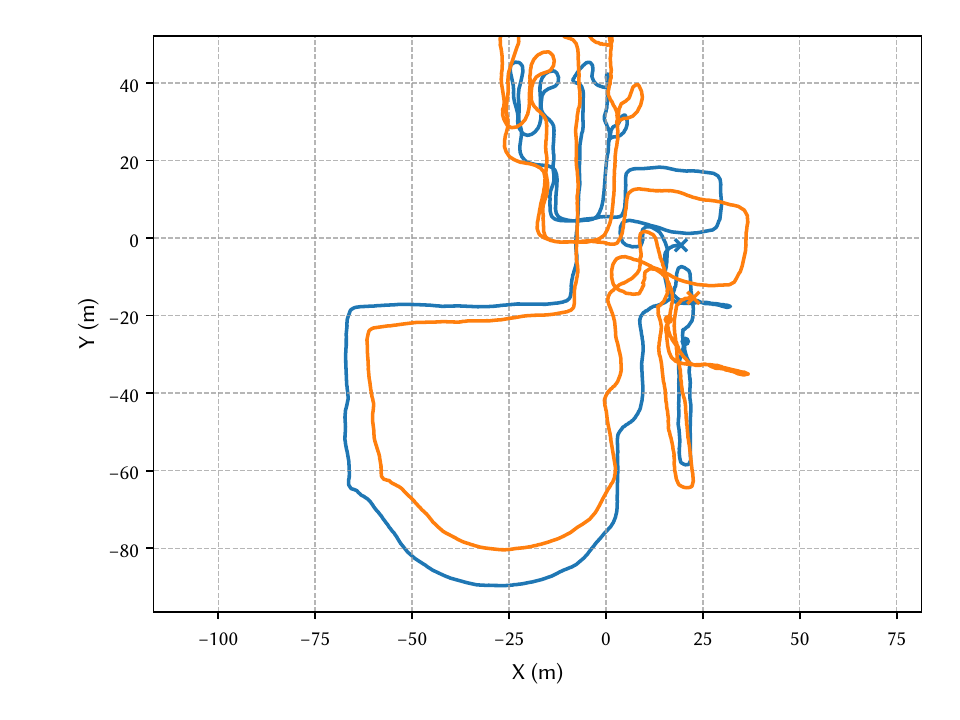}
      {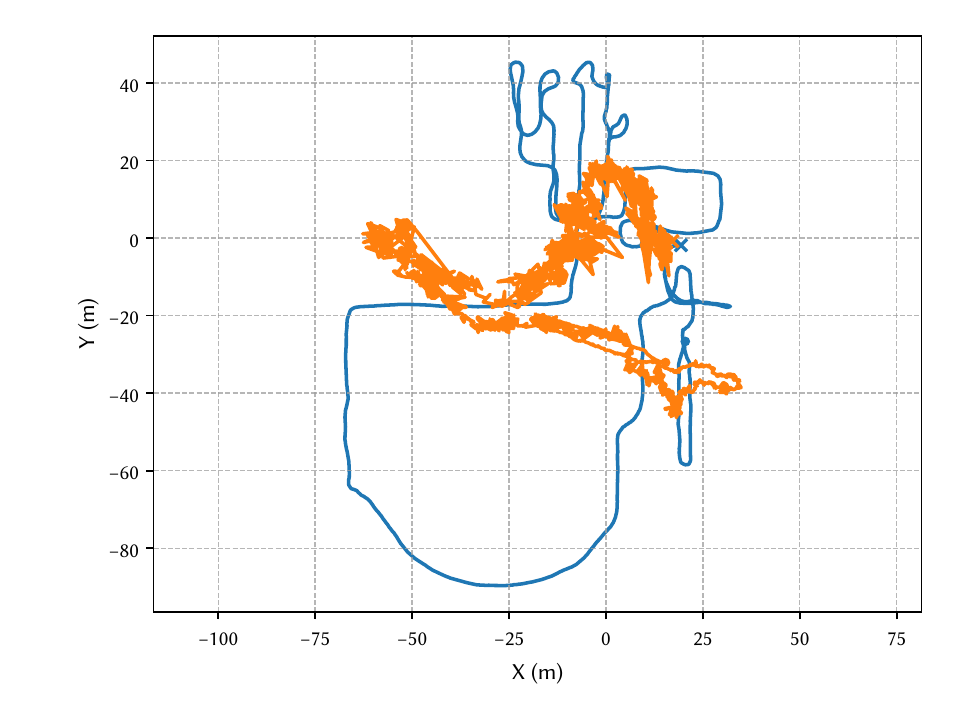}
      {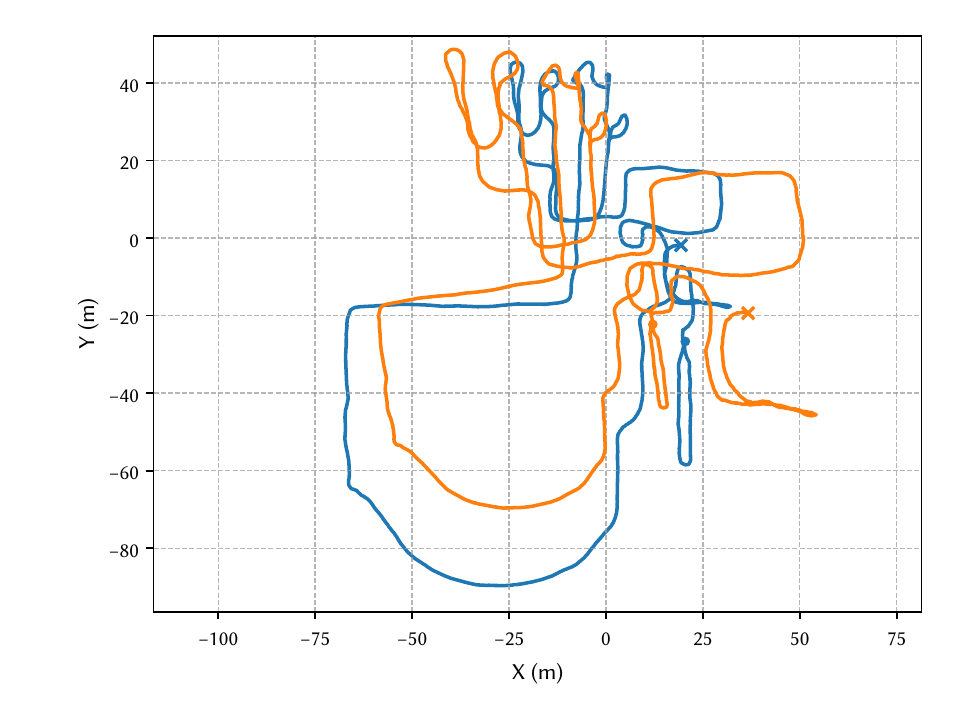}
      {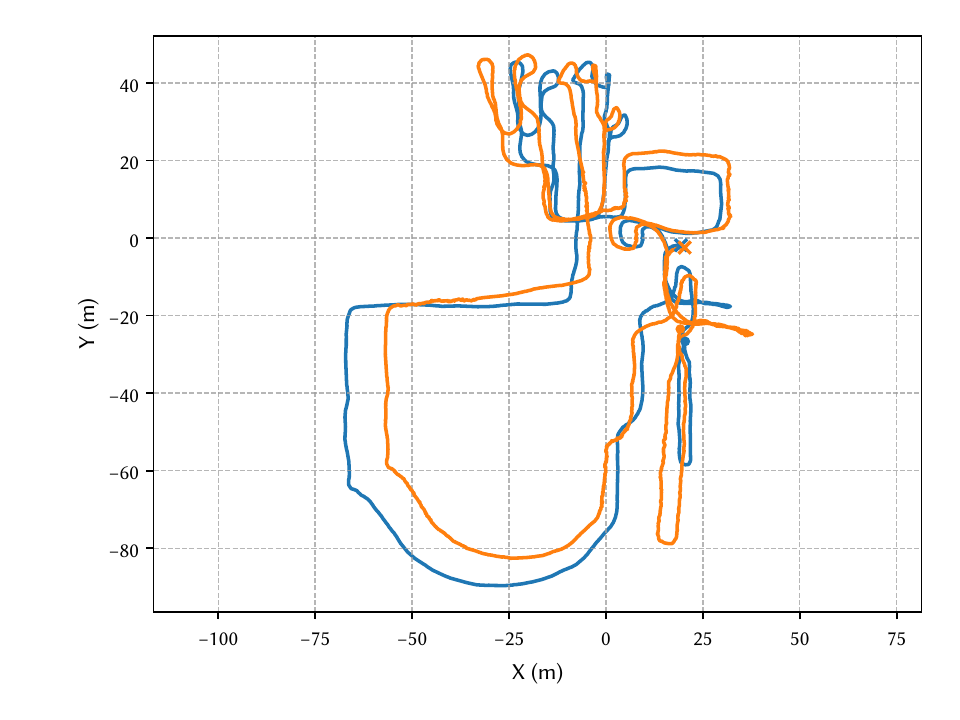}
  \end{tabular}%
  }
  \endgroup
  \captionsetup{skip=2pt}
  \caption{\textbf{Additional full-sequence trajectory comparisons.} Blue and orange denote the reference and estimated trajectories, respectively. Each estimate is aligned once over the full sequence by \(\mathrm{Sim}(3)\), matching the ATE protocol.}
  \label{fig:trajectory-comparison-supp}
\end{figure*}

\clearpage
\section{Interactive Qualitative Results}
\label{sec:interactive-qualitative-results}

\begin{figure}[htbp]
  \centering
  \begingroup
  \newcommand{\gamequalrow}[4]{%
    {\centering\small\textit{#1}: #2\par}%
    \vspace{3pt}%
    \noindent
    \includegraphics[width=0.468\linewidth]{#3}%
    \hfill
    \includegraphics[width=0.468\linewidth]{#4}%
    \par\vspace{8pt}%
  }
  \gamequalrow
    {GTA V}
    {Night-to-day transition}
    {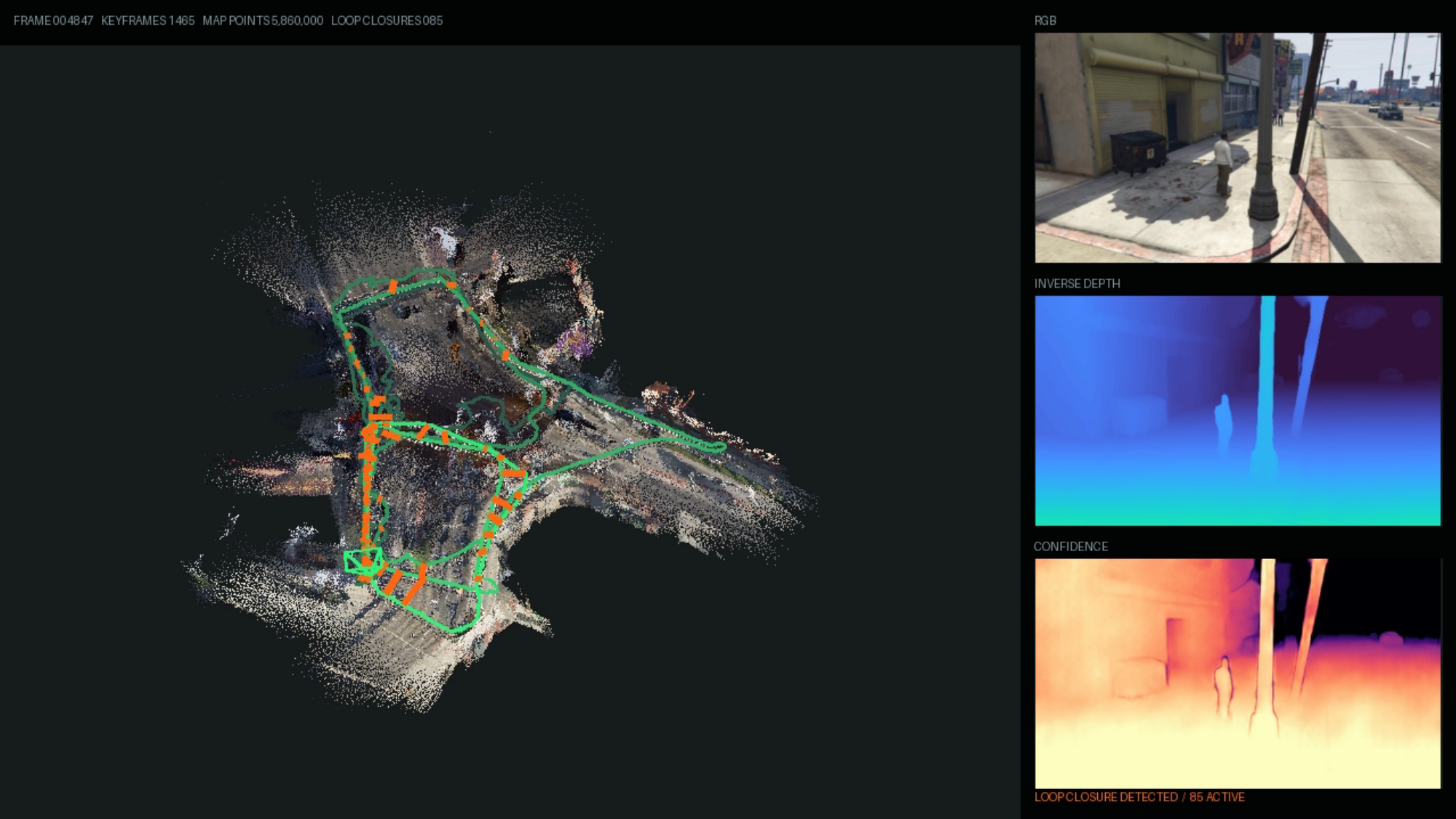}
    {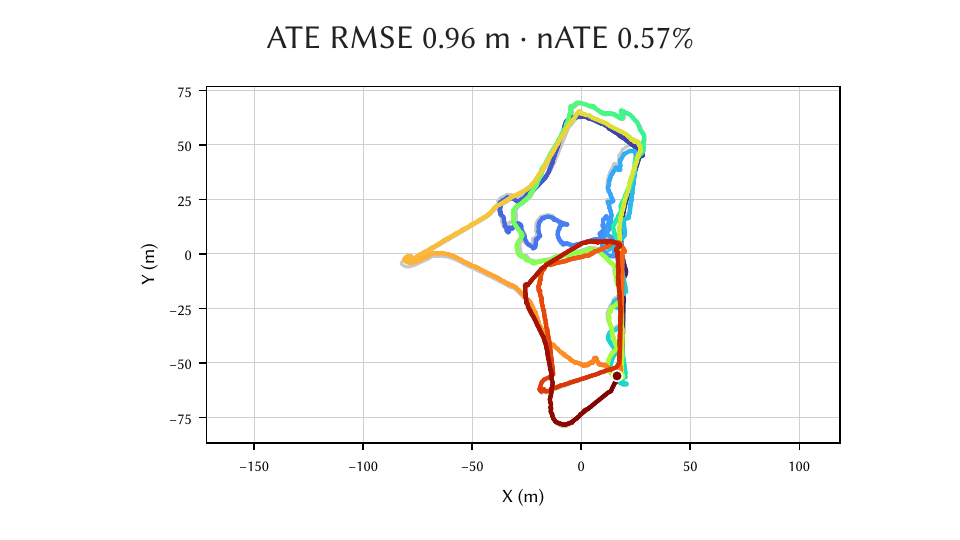}
  \gamequalrow
    {GTA V}
    {Abrupt camera-speed changes}
    {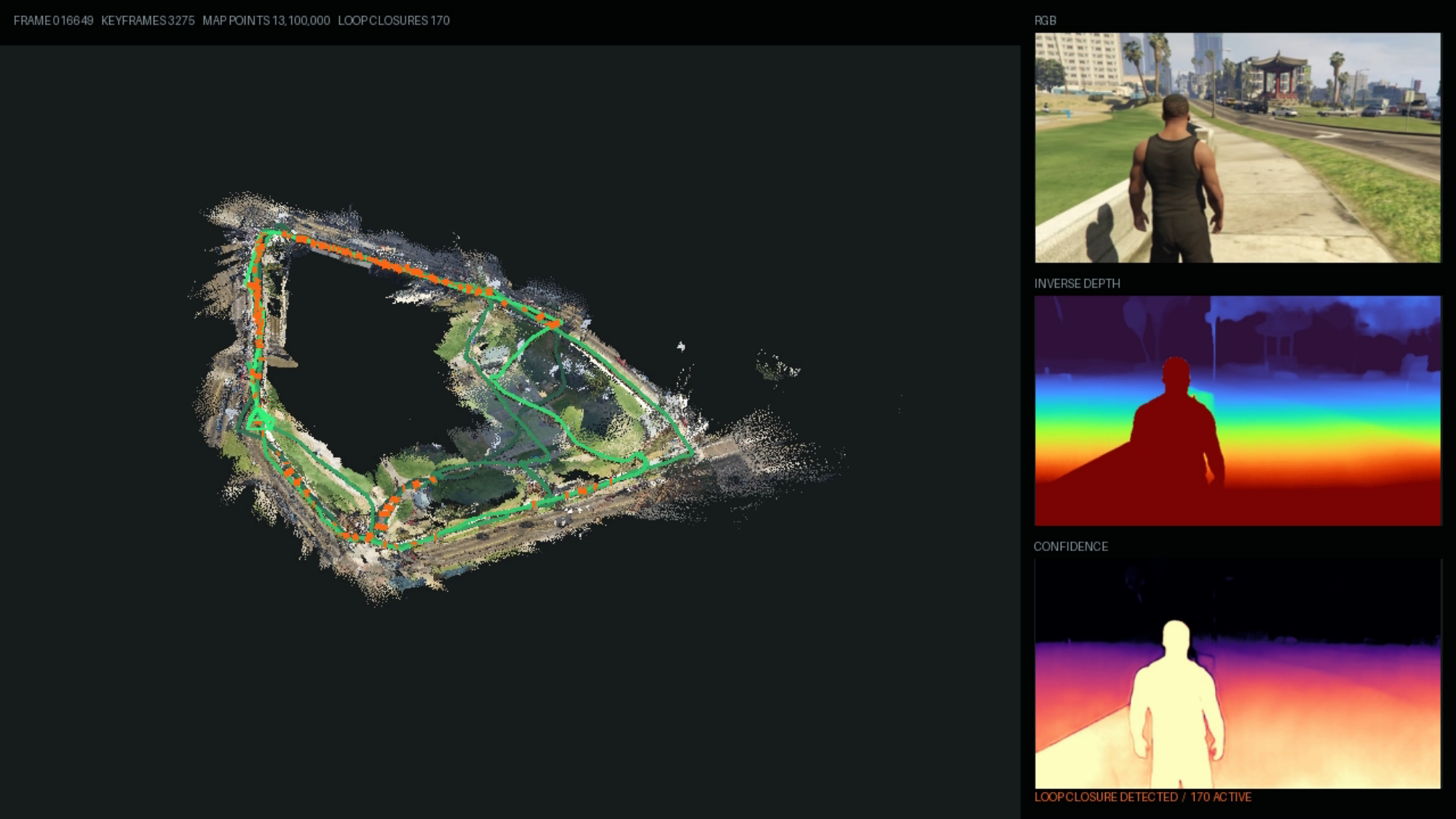}
    {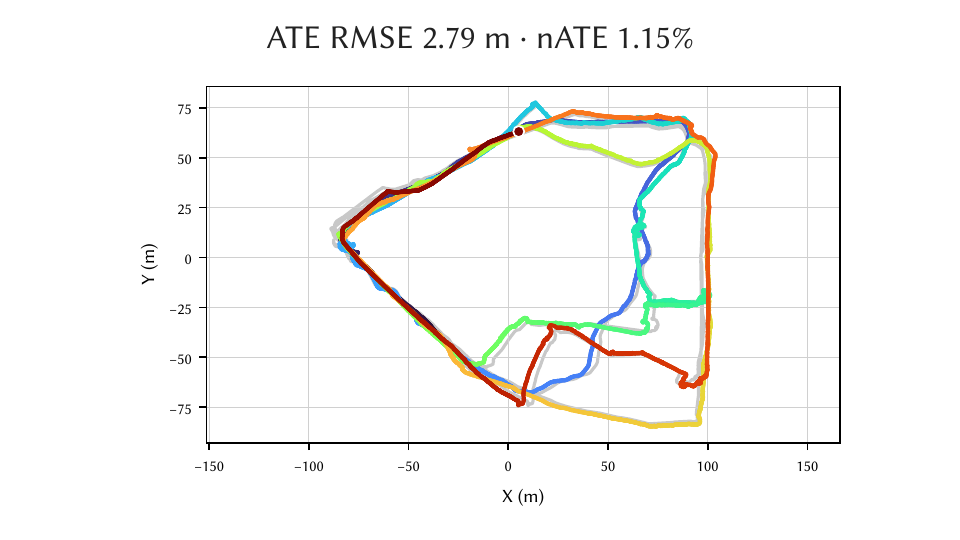}
  \gamequalrow
    {Assassin's Creed Origins}
    {Dense loop closures}
    {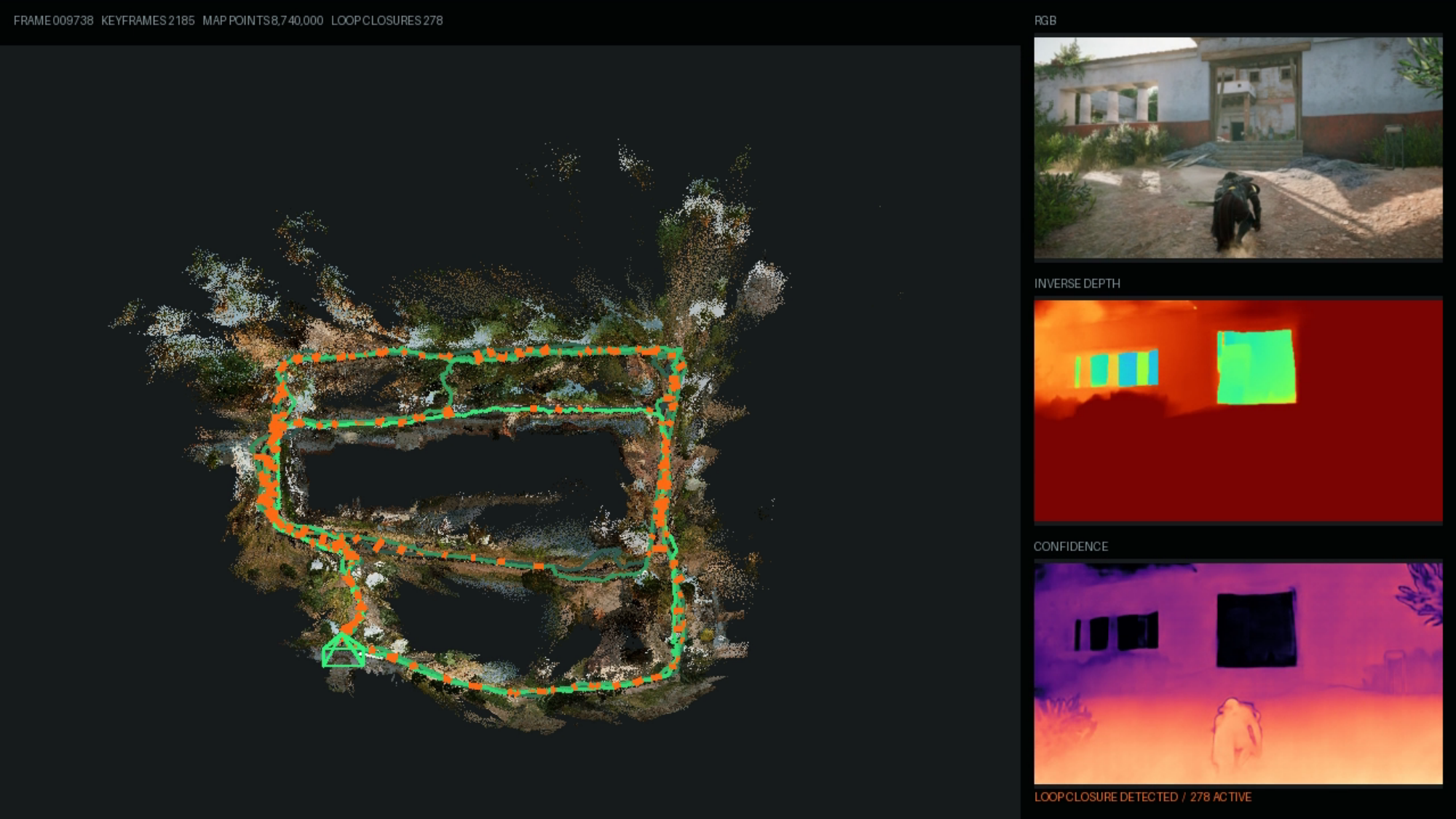}
    {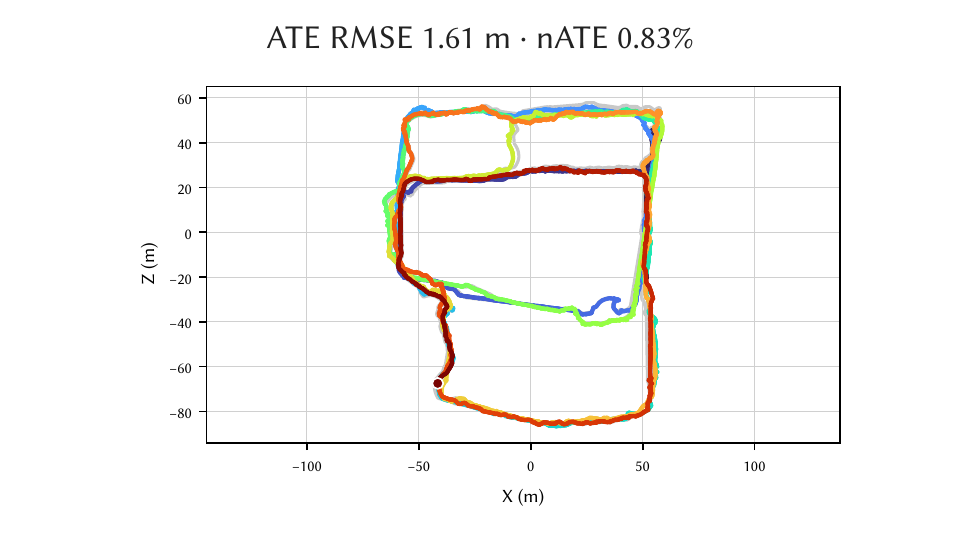}
  \endgroup
  \captionsetup{skip=2pt}
  \caption{\textbf{Results on three game sequences. }Each row pairs the
  terminal system visualization (left) with the final trajectory against
  ground truth (right). The trajectory panels report ATE RMSE and normalized
  ATE (nATE).The trajectories estimated by our system are color-coded from blue to red over time, while the ground-truth trajectories are shown in gray. As illustrated, our estimated trajectories closely align with the ground truth across nearly all sequences.}
  \label{fig:game-qualitative-supp}
\end{figure}

Figure~\ref{fig:game-qualitative-supp} shows qualitative SLAM results on 3 game sequences with different challenging conditions. The first GTAV sequence contains a transition \textbf{from nighttime to daytime}, while the second GTAV sequence features a \textbf{sudden change in speed} of camera motion. The third sequence, from Assassin's Creed Origins, contains \textbf{dense loop closures}. 

The complete videos are available on the \href{https://robbyant-research.github.io/Map2SLAM/}{\textcolor{magenta}{project page}} and provide a clearer view of the system's behavior over time than the static frames shown here. We therefore encourage readers to explore the project page for a more comprehensive visualization of these sequences and our SLAM results.

\end{document}